\documentclass[a4paper,fleqn]{cas-sc}

\usepackage[authoryear,longnamesfirst]{natbib}
\usepackage{rotating}
\usepackage{tabularx}
\usepackage{placeins}
\usepackage{arydshln}
\usepackage{enumitem}
\usepackage{xurl}

\setlist[enumerate,1]{leftmargin=1.25em,label=\alph*)}

\begin{document}
\let\WriteBookmarks\relax
\def\floatpagepagefraction{1}
\def\textpagefraction{.001}
\hypersetup{hidelinks}

\shorttitle{Do Schwartz Higher-Order Values Help Sentence-Level Human Value Detection?}
\shortauthors{Víctor Yeste and Paolo Rosso}

\title[mode=title]{Do Schwartz Higher-Order Values Help Sentence-Level Human Value Detection? A Study of Hierarchical Gating and Calibration}

\author[1,2]{Víctor Yeste}[
  orcid=0000-0002-3660-8347,
  twitter=victormyeste,
  linkedin=victoryeste
]
\cormark[1]
\ead{vicyesmo@upv.es}
\ead[url]{https://victoryeste.com/en/}
\credit{Conceptualization, Methodology, Software, Validation, Formal analysis, Investigation, Resources, Data curation, Writing - Original draft, Writing - Review \& Editing, Visualization, Project administration}

\author[1,3]{Paolo Rosso}[
  orcid=0000-0002-8922-1242
]
\ead{prosso@dsic.upv.es}
\credit{Supervision, Writing - Review \& Editing}

\affiliation[1]{organization={PRHLT Research Center, Universitat Politècnica de València},
            city={Valencia},
            postcode={46022},
            country={Spain}}

\affiliation[2]{organization={School of Science, Engineering and Design, Universidad Europea de Valencia},
            city={Valencia},
            postcode={46010},
            country={Spain}}

\affiliation[3]{organization={Valencian Graduate School and Research Network of Artificial Intelligence (ValgrAI)},
            city={Valencia},
            country={Spain}}

\cortext[1]{Corresponding author}

\begin{abstract}
Human value detection from single sentences is a sparse, imbalanced multi-label task. We study whether Schwartz higher-order (HO) categories help this setting on ValueEval'24 / ValuesML (74K English sentences) under a compute-frugal budget. Rather than proposing a new architecture, we compare direct supervised transformers, hard HO$\rightarrow$values pipelines, Presence$\rightarrow$HO$\rightarrow$values cascades, compact instruction-tuned large language models (LLMs), QLoRA, and low-cost upgrades such as threshold tuning and small ensembles. HO categories are learnable: the easiest bipolar pair, \emph{Growth vs.\ Self-Protection}, reaches Macro-$F_1=0.58$. The most reliable gains come from calibration and ensembling: threshold tuning improves \emph{Social Focus vs.\ Personal Focus} from $0.41$ to $0.57$ ($+0.16$), transformer soft voting lifts \emph{Growth} from $0.286$ to $0.303$, and a Transformer+LLM hybrid reaches $0.353$ on \emph{Self-Protection}. In contrast, hard hierarchical gating does not consistently improve the end task. Compact LLMs also underperform supervised encoders as stand-alone systems, although they sometimes add useful diversity in hybrid ensembles. Under this benchmark, the HO structure is more useful as an inductive bias than as a rigid routing rule.
\end{abstract}

\begin{highlights}
\item HO values are learnable, but performance varies widely across pairs.
\item Hard HO gating does not reliably improve end-task value detection.
\item Threshold tuning yields consistent gains under compute frugal settings.
\item Small ensembles give the most reliable improvements across HO slices.
\item Small LLMs lag alone but add diversity in cross family ensembles.
\end{highlights}


\maketitle

\section{Introduction}
\label{sec:introduction}

Human values are enduring guiding principles that shape what people consider important, desirable, or worth protecting \citep{Rokeach1973,Schwartz1992,BardiSchwartz2003}. Because values are often expressed implicitly in language, detecting them in text matters for computational social science and NLP tasks that analyze public discourse, persuasion, stance, framing, and argumentation at scale \citep{Lazer2009}. Recent surveys of computational morality and value modeling summarize key approaches (lexicon-based signals, supervised classifiers, LLM-centric methods) and emphasize persistent challenges such as contextual ambiguity and domain sensitivity \citep{Reinig2024}. Related work on moral language in political and social media discourse shows that moral/value cues are typically sparse, indirect, and context-dependent \citep{HaidtJoseph2004,JohnsonGoldwasser2018}.

\begin{figure}[htbp]
    \centering
    \includegraphics[width=\linewidth]{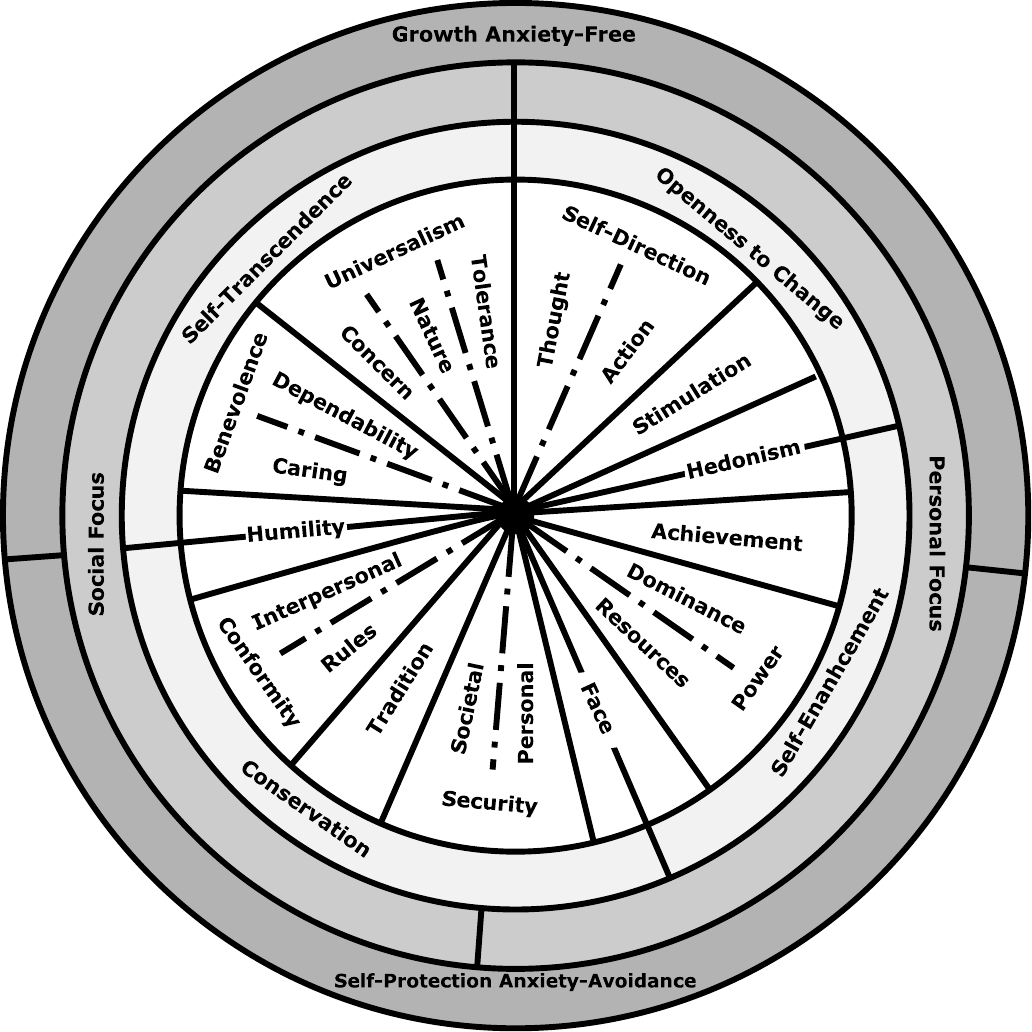}
    \caption{Schwartz's circular continuum of the 19 basic values. Adjacent values are compatible, whereas opposing values tend to conflict. Adapted from \citet{Schwartz2012}.}
    \label{fig:schwartz_continuum}
\end{figure}

Among social-science value frameworks, Schwartz's theory is widely adopted and empirically validated \citep{Schwartz1992}. The refined theory defines 19 basic values and groups them into higher-order (HO) categories (e.g., \emph{Openness to Change} vs.\ \emph{Conservation}) that capture compatibilities and conflicts \citep{Schwartz2012overview}. This hierarchy suggests a potential inductive bias for predicting fine-grained values when labels are sparse or ambiguous. Figure~\ref{fig:schwartz_continuum} shows the circular motivational continuum of the 19 values, where adjacent values are compatible and opposing values are in conflict \citep{Schwartz2012}.

Recent shared tasks operationalize value detection as sentence-level, multi-label prediction, enabling controlled comparisons \citep{Kiesel2023}. The Touché 2024 Human Value Detection task (ValueEval'24) is built on ValuesML, where sentences are annotated for which Schwartz values they express and whether a value is \emph{attained} or \emph{constrained} \citep{ValueEval24Zenodo,Touche2024}. Recent work highlights machine learning challenges and benchmark-driven evaluation, showing how dataset design and label distributions shape performance \citep{Rink2025}. These benchmarks reveal realistic difficulties: a sentence may express none, one, or many values; evidence is often implicit and lexically diffuse; and label prevalence is highly imbalanced \citep{Kiesel2023}. These properties strain standard multi-label pipelines and make calibration decisions (thresholds, probability reliability) especially important \citep{TsoumakasKatakis2007,Guo2017,SilvaFilho2023}.

In practical terms, reliable value detection could support several real-world uses. One is value-aware monitoring of political and advocacy messaging, where analysts may want to track how issues are framed in terms of security, tradition, autonomy, or concern for others across campaigns, debates, and media streams \citep{JohnsonGoldwasser2018}. A second is large-scale public-opinion and framing analysis, where value signals can help characterize how online communities or large text collections emphasize different normative priorities over time \citep{Borenstein2025,Rink2024}. A third is value auditing for AI-mediated communication: as LLMs increasingly generate or transform public-facing text, value detection can serve as a lightweight diagnostic for checking which values their outputs tend to foreground, suppress, or align with \citep{Yao2024,Shen2025,Ye2025}. Related recent work also uses LLMs to assess public-facing health media at scale, reinforcing the broader need for reliable automated analysis of consequential communication channels \citep{Zhou2026}.

A natural question is whether Schwartz's HO structure improves fine-grained sentence-level detection. Hierarchical classification can help by injecting structure, constraining hypotheses, and sharing statistical strength \citep{SillaFreitas2011}. But in noisy sentence-level settings, hard constraints can amplify upstream errors: if a parent prediction is uncertain, strict gating can suppress true positives and reduce recall on already sparse labels. This tension is tied to calibration, since hierarchical pipelines are sensitive to thresholds and probability miscalibration \citep{Valmadre2022}.

We address this tension through a controlled, compute-bounded empirical study of when HO categories help and how they should be used. Rather than proposing a new hierarchical learner, we compare: (i) direct multi-label prediction, (ii) a Category$\rightarrow$Values hierarchy that constrains fine-grained outputs, and (iii) a \emph{Presence$\rightarrow$Category$\rightarrow$Values} cascade that first filters sentences predicted to contain any value. We also test low-cost levers that often matter in practice, including threshold calibration and simple ensembling \citep{Wolpert1992,Breiman1996,FreundSchapire1997,Breiman2001}. Finally, we benchmark compact instruction-tuned LLMs under the same budget, motivated by evidence that prompting and instruction tuning can be competitive without task-specific architectural changes \citep{Brown2020,Ouyang2022,Chung2024}. The scope is deliberately benchmark-driven: the goal is to characterize behavior on ValueEval'24 / ValuesML under fixed compute, not to claim immediate generalization across domains, languages, or annotation schemes.

Operationally, this makes the study a set of component-wise ablations under fixed compute, where hierarchy injection, thresholding, auxiliary features, prompting/fine-tuning choices, and ensembling are varied while the evaluation protocol remains constant.

\paragraph{Research questions.}
We structure the study around the following research questions:

\begin{enumerate}[label=\textbf{RQ\arabic*.}, leftmargin=*, itemsep=0.25em]
  \item \textbf{Are HO values learnable from single sentences?}
  Can we reliably detect the eight Schwartz HO categories from a single sentence, and which compute-frugal signals and model families work best?

  \item \textbf{Do HO gates help downstream basic-value prediction?}
  Does inserting an HO category detector as a gate before predicting the 19 basic values improve out-of-sample Macro-$F_1$ compared to a single-stage \emph{Direct} model?

  \item \textbf{Does a Presence$\rightarrow$Category cascade improve over Category-only?}
  If we add a \emph{Presence} gate before the HO gate (\emph{Presence$\rightarrow$ \allowbreak Category$\rightarrow$Values}), does this hierarchy outperform (a) \emph{Direct} prediction and (b) \emph{Category$\rightarrow$Values} on the test set?

  \item \textbf{What low-cost knobs actually move the needle?}
  Across HO detection and the hierarchical pipeline, which lightweight signals (lexica, topic vectors, short local context) and which calibration/ensembling choices yield statistically supported gains under fixed compute?

  \item \textbf{Where do small LLMs fit?}
  Under the same budget, how do instruction-tuned $\le$10B LLMs (zero-shot, few-shot, and QLoRA) compare to supervised DeBERTa-based models for HO detection and for the hierarchy-driven pipeline?
\end{enumerate}

\paragraph{Working hypotheses.}
The framework is guided by three testable hypotheses. \textbf{H1:} HO categories should be easier to predict than fine-grained values because they aggregate multiple basic values into coarser labels, reducing sparsity. \textbf{H2:} hard hierarchical gating should improve structural consistency and may improve precision, but can reduce end-task recall through upstream false negatives. \textbf{H3:} under strong label imbalance, label-wise threshold tuning should be a more reliable source of gains than hard gating because it adjusts decision rules without introducing additional upstream failure points.

Our contributions are:
\begin{itemize}
    \item A careful, compute-bounded empirical study of whether Schwartz HO categories improve sentence-level value detection on ValueEval'24/ValuesML, including analyses by canonical bipolar HO pairs.
    \item A controlled comparison of HO-aware strategies (conditioning, hard gating, cascades) that identifies when hierarchy helps and when it fails due to error propagation.
    \item Evidence that, under this benchmark, calibration-aware thresholding and small ensembles are more reliable sources of gain than hard hierarchical routing.
\end{itemize}

Overall, we contribute a practical empirical characterization of when value structure helps under fixed compute, and show that \emph{hard} hierarchical constraints are brittle in this sentence-level setting, while calibration and small ensembles deliver the most reliable gains under this benchmark.

The rest of the paper is organized as follows. Section~\ref{sec:related-work} reviews prior work on value and moral language detection, benchmarks, hierarchical and multi-label learning, calibration/ensembling, and the use of transformers and instruction-tuned LLMs. Section~\ref{sec:method} describes the task, dataset, model variants, and compute-frugal protocol. Section~\ref{sec:results} reports results and analyzes when HO structure helps (or hurts). Section~\ref{sec:discussion} discusses implications and answers the research questions. Section~\ref{sec:conclusions-future-work} concludes and outlines future work. Tables and figures labeled Sx (e.g., Table S4) refer to the Supplementary Material.

\section{Related work}
\label{sec:related-work}

\subsection{Human values and moral frameworks in NLP}
Human values are commonly operationalized as relatively stable guiding principles that shape preferences and judgments \citep{Rokeach1973,Schwartz1992,BardiSchwartz2003}. For an NLP-oriented synthesis of how morality/value constructs are used in text analysis, see \citet{Reinig2024}. Among taxonomies, Schwartz's theory is especially attractive for NLP because it provides (i) a refined, fine-grained set of basic values and (ii) a principled HO organization that captures compatibilities and conflicts \citep{Schwartz2012}. Computational work on moral language often draws on Moral Foundations Theory (MFT) to study moral rhetoric in political and social discourse \citep{Graham2009}. Although MFT and Schwartz address different constructs, both highlight a key modeling challenge: moral/value signals in text are often indirect, sparse, and diffuse rather than explicitly labeled \citep{Graham2011,Haidt2012}. Beyond classification, NLP has also been used to elicit structured value representations for downstream systems, such as extracting value promotion schemes \citep{GarciaRodriguez2025}.

An extensive line of work explores feature- and lexicon-driven prediction as an interpretable, low-cost alternative to heavy end-to-end models \citep{Hopp2021,Hoover2020}. \citet{Araque2020} propose MoralStrength, which extends the Moral Foundations Dictionary with embedding-based similarity. \citet{GonzalezSantos2023} study moral foundations assignment in the movie domain using word embeddings and semantic similarity. These studies motivate our compute-frugal perspective: lightweight signals can be useful, but performance depends on how prior structure is injected.

Recent work expands the scope of value detection beyond classic lexicon settings to online community discourse and multimodal platforms, and proposes methods for large text collections. For example, value expressions are analyzed in online communities \citep{Borenstein2025} and in multimodal influencer content \citep{Starovolsky-Shitrit2025}, while context-dependent markup schemes are proposed for large-scale value/sentiment detection in social media corpora \citep{Rink2024}. In parallel, LLM-based value identification has emerged as a lightweight alternative for text-only settings \citep{Zhu2025}.

\subsection{Benchmarks and shared tasks for value detection}
The field has converged on shared benchmarks to enable controlled comparisons and expose realistic difficulty factors such as multi-label outputs, class imbalance, and cross-domain variation. For argumentation, \citet{Kiesel2022} introduce a value-annotated benchmark for identifying human values behind arguments, and the Touch{\'e}/ValueEval line of tasks systematizes evaluation and reporting \citep{Mirzakhmedova2024,Kiesel2023}. In Touch{\'e} 2024, the Human Value Detection task frames value detection as sentence-level prediction under operational constraints typical of applied NLP pipelines \citep{ValueEval24Zenodo,Touche2024}.

Within this context, recent systems emphasize cascaded decision processes and threshold control under label sparsity. The best-performing English system reported for Touch{\'e}/CLEF 2024 uses a cascade to structure decisions and reduce spurious positives \citep{Yeste2024}. \cite{Yeste2026} study sentence-level value detection with \emph{moral presence} gating and compute-frugal transformer ensembles, providing a strong baseline that we extend by focusing on Schwartz HO categories and their use as hierarchical structure.

Beyond shared tasks, recent resource suites aim for broader coverage across frameworks and domains; for example, MoVa aggregates multiple labeled datasets and benchmarks across moral/value theories to enable more generalizable evaluation \citep{Chen2025}.

\subsection{Hierarchical structure and multi-label learning for value prediction}
Our core question---whether HO structure helps basic-value prediction---connects to two mature ML literatures: multi-label learning and hierarchical classification. Multi-label classification is difficult when labels are non-mutually exclusive, skewed, and supported by limited positive evidence \citep{ZhangZhou2014}. Hierarchical classification studies how label taxonomies can share statistical strength and impose structure. Recent work on hierarchical text classification explores label-based attention and global label-graph modeling to share information across levels, mitigate error propagation, and improve lower-level labels under hierarchical imbalance \citep{Zhang2022,Liu2024}. These insights motivate our HO-aware variants: HO labels may reduce effective sparsity, but rigid enforcement can increase error propagation when parent predictions are uncertain \citep{SillaFreitas2011,Wang2024}.

A related theme is the role of \emph{context} when a single sentence provides limited evidence. Hierarchical document models (e.g., sentence-then-document encoders) improve classification by modeling multi-granular context \citep{Yang2016}, and prior analyses in hierarchical text classification show that representation choices across levels can materially affect downstream performance \citep{Stein2019}. This suggests HO structure may be most effective as guidance combined with careful control of context, rather than as a brittle hard constraint.

\subsection{Calibration, thresholding, and ensemble robustness under imbalance}
Because value detection is multi-label and imbalanced, performance is often dominated by decision rules that map scores to binary labels. Classical calibration work (e.g., Platt scaling; \citet{Platt1999}) and later studies on probability reliability show how miscalibration can distort precision/recall trade-offs, especially with per-label thresholds \citep{SilvaFilho2023}. This motivates our emphasis on threshold tuning as a compute-frugal but high-leverage component.

Ensembling is another long-standing route to robustness under limited data and noisy supervision \citep{Dietterich2000,Rokach2010}. In value detection, different models can capture complementary cues, so small ensembles often deliver consistent gains without increasing single-model capacity. This aligns with recent dynamic ensemble work for multi-label classification under label dependence and imbalance \citep{Zhu2023}. We build on this principle and the compute-frugal ensemble methodology in \citet{Yeste2026} to test whether HO-aware modeling improves beyond calibration and modest diversity. More specifically, whereas prior work mainly motivates calibration and ensembling as generally useful tools under imbalance, our contribution is to contrast them directly with hard hierarchical routing under a fixed benchmark and compute budget, showing that the former are the more reliable source of gains in this setting.

\subsection{Transformers and instruction-tuned LLMs for moral/value classification}
Modern value detection systems typically rely on transformer encoders fine-tuned for classification \citep{Devlin2019,He2021}. Instruction-tuned LLMs have popularized prompt-based classification as a lightweight alternative that avoids task-specific architectural changes \citep{Zhao2021,Wei2022,Liu2023}. Work comparing prompting and supervised adaptation for human values motivates treating \emph{prompting vs.\ fine-tuning} as an explicit design choice \citep{Sun2024}. For sentence-level, sparse multi-label settings, prompt-based LLMs still face calibration and recall challenges, often with less control over score distributions. Parameter-efficient methods such as QLoRA offer a middle ground between pure prompting and full fine-tuning \citep{Hu2022,Dettmers2024}. Recent work also studies deployable generative-AI systems beyond generic prompting, including small language models for constrained environments and deterministic LLM-based assessment pipelines \citep{Nunez2026,Zhou2026}.

Recent LLM work usefully splits into two strands. The first is closest to our task: \emph{value detection/classification from text}, where LLMs are used to assign value labels to inputs. Recent examples include EAVIT, which studies LLM-based human value identification from text, and MoVa, which targets more generalizable morals/value classification across a broader aggregated benchmark suite \citep{Zhu2025,Chen2025}. The second studies \emph{value alignment/representation in the models themselves}, asking which values LLMs express, prioritize, or align with rather than predicting values in input text. Examples include Value FULCRA, which maps LLM outputs to Schwartz value dimensions \citep{Yao2024}, UniVaR, which learns a high-dimensional representation of value distributions in LLMs across models and languages \citep{Cahyawijaya2025}, ValueCompass, which measures contextual alignment between human and LLM values across scenarios \citep{Shen2025}, and psychometric-style measurement of human/LLM values from text \citep{Ye2025}. Recent analyses further probe value consistency and cultural alignment in LLMs \citep{Rozen2025,Segerer2025,Biedma2024}. Our experiments are positioned in the first strand, while the second mainly motivates why value-aware analysis of LLM outputs matters.

Taken together, these threads motivate the design space evaluated in this paper. We adopt the shared-task framing and compute-frugal discipline established in \citet{ValueEval24Zenodo,Touche2024,Yeste2026}, and we focus the comparison on \emph{how} HO structure is injected (conditioning vs.\ hard gating/cascades) relative to strong, practically motivated baselines based on calibration and small ensembles. This sets up the methodological choices introduced next (Section~\ref{sec:method}).

\section{Methodology}
\label{sec:method}

\subsection{Problem formulation and label spaces}
\label{sec:method_problem}

We study sentence-level human value detection under Schwartz's refined theory \citep{Schwartz2012overview}. Each sentence may express none, one, or multiple values. Let $s$ be a sentence and $\mathbf{y}^{(19)}(s) \in \{0,1\}^{19}$ the binary vector over the 19 basic values.\footnote{The benchmark provides \emph{attained} and \emph{constrained} annotations per value; we collapse them into a single \emph{expressed value} signal (Section~\ref{sec:method_data}).}

\paragraph{HO categories.}
To test whether coarser abstractions help, we deterministically derive eight HO binary labels from the 19 values following \citet{Schwartz2012overview}: \emph{Openness to Change}, \emph{Conservation}, \emph{Personal Focus}, \emph{Social Focus}, \emph{Self-Enhancement}, \emph{Self-Transcendence}, \emph{Growth}, and \emph{Self-Protection}. Let $\mathcal{C}$ be the set of HO categories and $\mathcal{V}_c \subseteq \{1,\dots,19\}$ the basic values grouped under $c \in \mathcal{C}$. We define:
\begin{equation}
y^{(\mathrm{HO})}_c(s) = \mathbb{I}\Bigl[\exists\, v \in \mathcal{V}_c:\ y^{(19)}_v(s)=1\Bigr],
\label{eq:ho_or}
\end{equation}
This yields $\mathbf{y}^{(\mathrm{HO})}(s)\in\{0,1\}^{8}$. The value-to-HO mapping is fixed by theory and reported in \ref{app:ho-mapping}.

\paragraph{Hierarchy representation and consistency.}
Let $M \in \{0,1\}^{|\mathcal{C}|\times 19}$ be the binary incidence matrix with
\begin{equation}
M_{cv} =
\begin{cases}
1, & \text{if } v \in \mathcal{V}_c,\\
0, & \text{otherwise.}
\end{cases}
\label{eq:incidence_matrix}
\end{equation}
Then Eq.~\eqref{eq:ho_or} can be written compactly as
\begin{equation}
\mathbf{y}^{(\mathrm{HO})}(s) = \mathbb{I}\bigl[M \mathbf{y}^{(19)}(s) \ge \mathbf{1}\bigr],
\label{eq:ho_matrix}
\end{equation}
where the inequality is interpreted element-wise. A prediction pair $\bigl(\hat{\mathbf{y}}^{(\mathrm{HO})}(s), \hat{\mathbf{y}}^{(19)}(s)\bigr)$ is \emph{hierarchy-consistent} if, for every value $v$ and HO category $c$, $M_{cv}=1$ and $\hat{y}^{(19)}_v(s)=1$ imply $\hat{y}^{(\mathrm{HO})}_c(s)=1$. This formalizes the structural constraint that a basic value cannot be predicted present while all of its parent HO categories are predicted absent.

\paragraph{Presence label.}
We also consider a binary \emph{Presence} gate that flags whether any value is expressed:
\begin{equation}
y^{(\mathrm{Pres})}(s) = \mathbb{I}\Bigl[\sum_{v=1}^{19} y^{(19)}_{v}(s) > 0\Bigr].
\label{eq:presence}
\end{equation}

\paragraph{Bipolar evaluation slices.}
In addition to performance over all eight HO labels, we analyze the four canonical bipolar pairs to expose asymmetries in learnability and error propagation: (i) \emph{Openness to Change} vs.\ \emph{Conservation}, (ii) \emph{Self-Enhancement} vs.\ \emph{Self-Transcendence}, (iii) \emph{Personal Focus} vs.\ \emph{Social Focus}, and (iv) \emph{Growth} vs.\ \emph{Self-Protection} \citep{Schwartz2012overview}. For each pair, we report Macro-$F_1$ averaged over the two poles.

\begin{figure}[htbp]
    \centering
    \includegraphics[width=\linewidth]{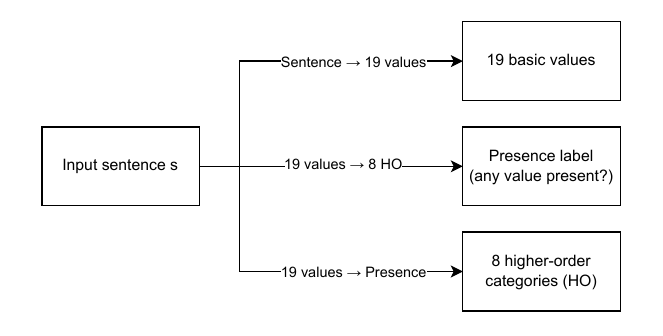}
    \caption{Sentence-level prediction setup and label spaces: 19 basic values, 8 HO categories derived by OR-ing values within each group (Eq.~\ref{eq:ho_or}), and a binary \emph{Presence} label (Eq.~\ref{eq:presence}).}
    \label{fig:task_labelspace}
\end{figure}

Figure~\ref{fig:task_labelspace} summarizes the relationship between the sentence input and the three label spaces.

\subsection{Dataset and preprocessing}
\label{sec:method_data}

We use the official train/validation/test split \citep{ValueEval24Zenodo,Touche2024} at \emph{sentence level}. To keep modeling choices unified and compute controlled, we use the benchmark's English version (machine-translated sentences provided by the benchmark). This yields $74{,}231$ sentences: $44{,}758$ train, $14{,}904$ validation, and $14{,}569$ test. The split is at the \emph{text} level; all sentences inherit their source text's split.

Detailed label prevalence statistics (basic values and derived HO categories) for each split are reported in \ref{app:prevalence}.

\paragraph{Label construction.}
For each of the 19 values, the benchmark provides \emph{attained} and \emph{constrained} signals with values in $\{0, 0.5, 1\}$, where $0.5$ denotes \emph{unclear}. We binarize by treating any non-zero annotation as evidence of expression and collapse attained/constrained into a single label:
that is, a value is marked as expressed if either its attained or constrained signal is non-zero.

\subsection{Model families}
\label{sec:method_models}

We study three compute-frugal model families under a common protocol: (i) supervised transformer encoders, (ii) instruction-tuned LLMs used via prompting (zero-/few-shot), and (iii) parameter-efficient LLM fine-tuning (QLoRA). Predictions are evaluated as multi-label decisions over either the 19 values or the eight HO categories.

\subsubsection{Direct multi-label prediction (supervised encoder)}
\label{sec:method_direct}

The \emph{Direct} approach follows \citet{Yeste2026}. Given the pooled sentence representation $\mathbf{h}(s)\in\mathbb{R}^{d}$, we apply a linear layer to produce logits $\mathbf{z}$ and probabilities $\hat{\mathbf{y}}=\sigma(\mathbf{z})$. Training minimizes standard multi-label binary cross-entropy:
we minimize standard multi-label binary cross-entropy over $K\in\{8,19\}$ labels. We do not use class weights, matching \citet{Yeste2026}.

\subsubsection{Category$\rightarrow$Values hierarchy (HO gating)}
\label{sec:method_cat_values}

To test whether HO structure acts as an inductive bias, we implement a two-stage pipeline:
\begin{enumerate}
  \item \textbf{Category stage}: predict $\hat{\mathbf{y}}^{(\mathrm{HO})}(s)$ over the eight HO categories.
  \item \textbf{Values stage}: predict the $19$ basic values, but condition decisions on the category predictions.
\end{enumerate}

We implement conditioning as a \emph{hard mask} that constrains which basic values can be positive. If value $v$ belongs to HO category $c(v)$, we set:
\begin{equation}
\hat{y}^{(19)}_{v}(s) \leftarrow \hat{y}^{(19)}_{v}(s)\cdot
\mathbb{I}\bigl[\hat{y}^{(\mathrm{HO})}_{c(v)}(s)\ge \tau_{c(v)}\bigr],
\label{eq:cat_mask}
\end{equation}
where $\tau_{c}$ is the tuned threshold for category $c$ (Section~\ref{sec:method_thresholds}). This makes a value permissive only when its parent HO category is predicted present.

\paragraph{Why hard gating can help or hurt.}
Hard gating enforces hierarchy consistency by construction, but it also introduces a recall bottleneck. Let $\mathrm{TP}_v^{\mathrm{dir}}$ denote the event that the ungated value model predicts value $v$ correctly on a sentence where $y_v^{(19)}(s)=1$, and let $\mathrm{TP}_{c(v)}^{\mathrm{HO}}$ denote the event that the parent HO gate is open on that sentence. Under Eq.~\eqref{eq:cat_mask}, a gated true positive for value $v$ requires both events to occur, so
\begin{equation}
\mathrm{Recall}_{\mathrm{gated}}(v)
=
\Pr\!\bigl(\mathrm{TP}_v^{\mathrm{dir}} \cap \mathrm{TP}_{c(v)}^{\mathrm{HO}} \mid y_v^{(19)}=1\bigr)
\le
\Pr\!\bigl(\mathrm{TP}_{c(v)}^{\mathrm{HO}} \mid y_v^{(19)}=1\bigr).
\label{eq:gated_recall_bound_simple}
\end{equation}
More conservatively,
\begin{equation}
\mathrm{Recall}_{\mathrm{gated}}(v)
\le
\mathrm{Recall}_{\mathrm{dir}}(v)\,
\Pr\!\bigl(\mathrm{TP}_{c(v)}^{\mathrm{HO}} \mid \mathrm{TP}_v^{\mathrm{dir}}, y_v^{(19)}=1\bigr)
\le
\mathrm{Recall}_{\mathrm{dir}}(v).
\label{eq:gated_recall_bound}
\end{equation}
Thus, hard gating can improve precision by eliminating structurally inconsistent positives, but it cannot create recall that was absent upstream and may reduce it whenever the parent HO detector misses a true case.

\subsubsection{Presence$\rightarrow$Category$\rightarrow$Values cascade}
\label{sec:method_presence_cat_values}

We test a three-stage cascade where \emph{Presence} acts as a first gate. The \emph{Presence} formulation follows \citet{Yeste2026}; here we include the \emph{Presence$\rightarrow$HO$\rightarrow$Values} cascade as one of the hierarchy-injection strategies under study:
\begin{enumerate}
  \item \textbf{Presence stage}: predict $\hat{y}^{(\mathrm{Pres})}(s)$.
  \item \textbf{Category stage}: if \emph{Presence} is positive, predict $\hat{\mathbf{y}}^{(\mathrm{HO})}(s)$; otherwise, output zeros for all HO labels.
  \item \textbf{Values stage}: apply the category-conditioned procedure from Eq.~\eqref{eq:cat_mask}.
\end{enumerate}
This cascade can improve precision by suppressing spurious positives on non-value sentences, but it can compound errors across stages. In particular, a correct final value prediction requires the conjunction of three events: a correct \emph{Presence} decision, a correct HO gate decision, and a correct value decision. Therefore, each additional hard gate narrows the feasible prediction space while introducing another potential source of false negatives. We quantify overall and per-pair effects in Section~\ref{sec:method_eval}.

\subsubsection{Instruction-tuned LLM baselines (definition prompting)}
\label{sec:method_llms}

We benchmark instruction-tuned open LLMs that fit on a single 8\,GB GPU (Llama~3.1~8B, Ministral~8B~2410, Qwen~2.5~7B, Gemma~2 9B). We use the \emph{definition-style} prompt (best in \citet{Yeste2026}), presenting one-line definitions for the 19 values from \citet{Schwartz2012overview}. The model returns \emph{only} a JSON array of applicable value names.

We evaluate zero-shot and few-shot prompting with $k\in\{1,2,4,8,16,\allowbreak20\}$ in-context examples. Few-shot templates prepend $k$ exemplars in the same schema and include at least one null exemplar (empty array) when $k=20$. We use greedy decoding with \texttt{max\_new\_tokens}=200. The definition-style template was inherited from the controlled prompt comparison in \citet{Yeste2026}, where it was the strongest prompt family among the variants tested, so we do not repeat a full prompt-engineering sweep here. The reported LLM results should therefore be read as a comparison under a restricted prompt budget chosen for compute parity and reproducibility, not as an upper bound on achievable LLM performance.

\paragraph{Post-processing and label mapping.}
LLM outputs are parsed as JSON arrays and mapped to a multi-hot vector $\hat{\mathbf{y}}^{(19)}(s)\in\{0,1\}^{19}$ by exact string matching. Invalid generations (non-JSON or out-of-vocabulary labels) are treated as empty predictions. For HO evaluation, we derive $\hat{\mathbf{y}}^{(\mathrm{HO})}(s)$ from $\hat{\mathbf{y}}^{(19)}(s)$ using Eq.~\eqref{eq:ho_or} to keep model families comparable.

\subsubsection{QLoRA fine-tuning (parameter-efficient LLM adaptation)}
\label{sec:method_qlora}

We additionally evaluate supervised QLoRA \citep{Dettmers2024}, following \citet{Yeste2026}. Based on validation screening, we fine-tune Gemma~2~9B as the backbone. We train only low-rank adapters (base model frozen) with learning rate $2\times10^{-4}$ and save only adapter weights.

We train two QLoRA variants: (i) \emph{QLoRA direct}, predicting the 19 values with rank $r{=}16$ and $\alpha{=}32$, three epochs, gradient accumulation 8, cosine schedule, and max length 512; and (ii) \emph{QLoRA hier}, with rank $r{=}8$, $\alpha{=}16$, three epochs, and max length 256. In both cases, LoRA targets the attention projection modules $\{\texttt{q\_proj,k\_proj,v\_proj,o\_proj}\}$.

For QLoRA models that output probabilities, thresholds are tuned on validation under the same protocol as supervised encoders (Section~\ref{sec:method_thresholds}) and then frozen for test.

\begin{figure}[htbp]
    \centering
    \includegraphics[width=0.75\linewidth]{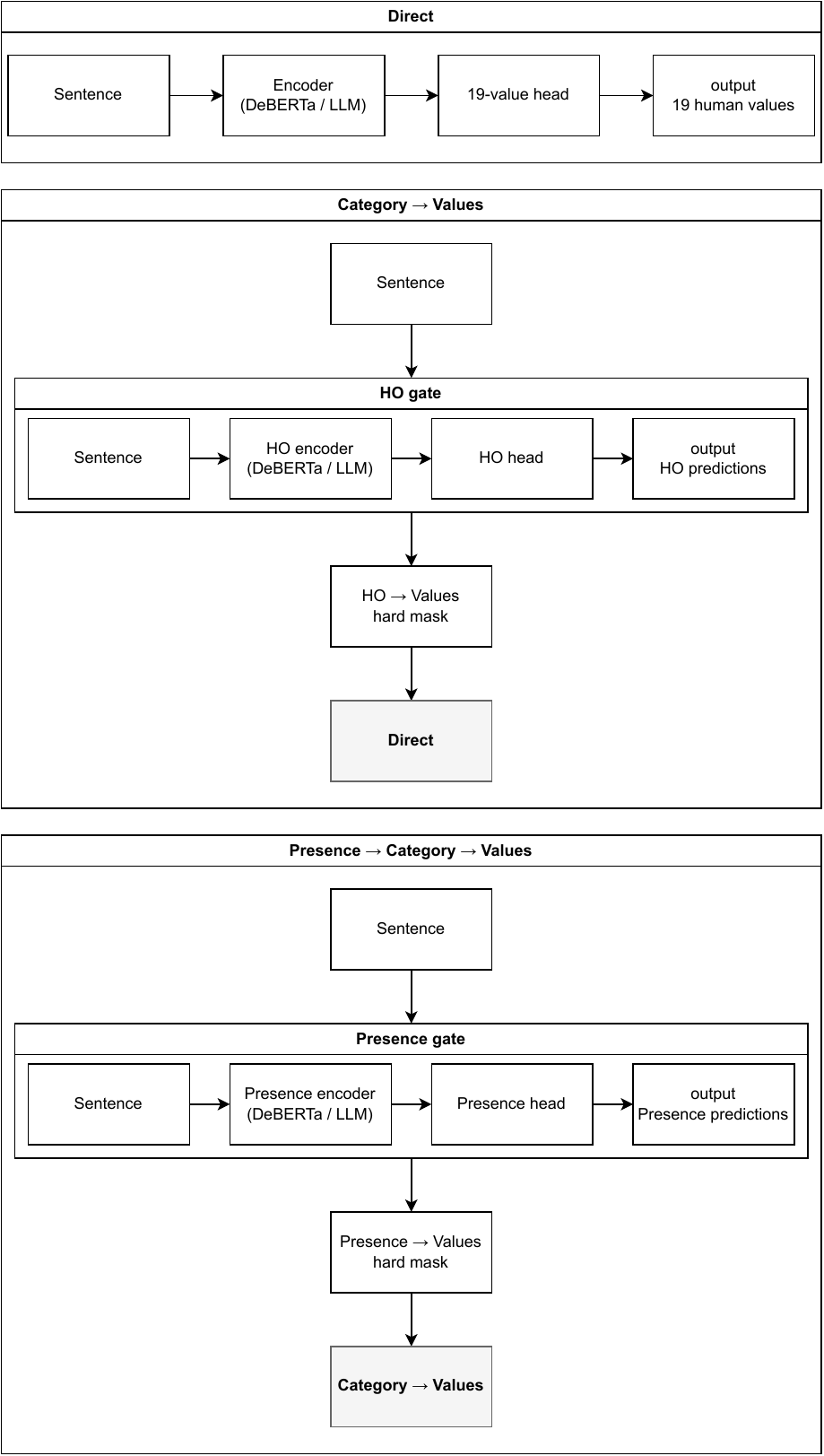}
    \caption{Schematic overview of the main model variants.}
    \label{fig:model_variants}
\end{figure}

Figure~\ref{fig:model_variants} summarizes the three main variants: a single-stage \emph{Direct} predictor, a two-stage \emph{Category$\rightarrow$Values} hierarchy with HO hard masks, and a three-stage \emph{Presence$\rightarrow$Category$\rightarrow$Values} cascade where a Presence gate filters sentences before HO and value prediction. All variants share the same encoder family and differ only in decision structure.

\subsection{Compute-frugal auxiliary signals}
\label{sec:method_features}

Beyond the plain transformer baseline, we evaluate compute-frugal add-ons that are inexpensive relative to the encoder forward pass and fit within an 8\,GB GPU budget:

\begin{itemize}
  \item \textbf{Short local context.} We concatenate up to the two previous sentences from the same source text to the current sentence (order preserved, separated by the model's separator token, e.g., \texttt{[SEP]}), then truncate to length 512. We also attach a $2\times|\mathcal{V}|$ vector encoding their value labels (gold at training; model predictions at validation/test in an auto-regressive manner). This vector is projected to a low-dimensional embedding and concatenated with the text representation to provide short-range discourse cues.

  \item \textbf{Lexicon-derived features.} We build sentence vectors from psycholinguistic, affective, moral, and value resources: LIWC-22 \citep{Boyd2022}, eMFD \citep{Hopp2021}, the Schwartz value lexicon in ValuesML \citep{Kiesel2023}, and affective lexica including NRC VAD \citep{Warriner2013,Mohammad2018}, NRC EmoLex \citep{MohammadTurney2013}, NRC Emotion Intensity \citep{MohammadKiritchenko2018}, and WorryWords \citep{Mohammad2024}. We aggregate token signals into sentence statistics (counts, relative frequencies, averaged intensities) and standardize using training-set statistics.

  \item \textbf{Topic features.} We attach topic-mixture vectors from unsupervised topic models trained on the training split only: LDA \citep{Blei200}, NMF \citep{LeeSeung1999}, and BERTopic \citep{Grootendorst2022}. At inference, validation/test sentences are mapped to topic vectors using the fixed models.
\end{itemize}

When auxiliary features are enabled (supervised encoders), we concatenate them to the pooled transformer representation before classification. Unless stated otherwise, features are computed from the same input sentence (plus optional short local context).

\subsection{Training protocol and compute parity}
\label{sec:method_training}

\paragraph{Experimental setup at a glance.}
All experiments use the official English ValueEval'24/ValuesML train/validation/test split at sentence level. We compare three model families under the same compute budget: supervised DeBERTa-base encoders, prompted instruction-tuned LLMs, and QLoRA-adapted LLMs. For supervised runs, we keep the backbone, maximum sequence length, and optimization protocol fixed across variants, and vary only the decision structure (\emph{Direct}, \emph{Category$\rightarrow$Values}, \emph{Presence$\rightarrow$Category$\rightarrow$Values}) and optional low-cost features. Model selection is performed on the validation split only, including threshold tuning and ensemble construction, and all final results are reported once on the held-out test split using the frozen validation-selected configuration.

Unless otherwise stated, we follow the supervised protocol of \citet{Yeste2026}. All transformer models fine-tune \texttt{microsoft/\allowbreak deberta-base} \citep{He2021} with a linear multi-label head. Inputs are tokenized and truncated/padded to length 512. We optimize with AdamW \citep{LoshchilovHutter2018} and a linear schedule with warmup, using batch size 4, gradient accumulation 4 (effective batch size 16), learning rate $2\times 10^{-5}$, weight decay 0.15, and up to 10 epochs with early stopping on validation Macro-$F_1$ (patience 4). Dropout is 0.1.

To keep comparisons fair and compute-frugal, we fix the encoder backbone and max sequence length across supervised variants, and select model variants and thresholds on validation only. All runs fit within a single 8\,GB GPU. For LLM prompting, we use quantized decoding where applicable; for QLoRA we fine-tune adapter weights only.

To keep the study compute-frugal while covering many strategies (direct prediction, hard gates/cascades, auxiliary signals, prompting, QLoRA, ensembles), we fix a single random seed (as in \citet{Yeste2026}) for supervised runs. This prioritizes breadth under a fixed budget and keeps differences tied to modeling choices rather than to random initialization. Instead of multiple seeds, we emphasize \emph{paired} evaluation: nonparametric sentence-level bootstrap uncertainty for $\Delta$Macro-$F_1$ and per-label paired tests (McNemar with FDR correction). We interpret differences below roughly 1--2 Macro-$F_1$ points conservatively and focus on effects supported by paired tests.

\subsection{Threshold calibration}
\label{sec:method_thresholds}

We map predicted probabilities to binary decisions using thresholds tuned on validation and then frozen for test. Let $\hat{p}_k(s)$ be the predicted probability for label $k$ and $\hat{y}_k(s)=\mathbb{I}[\hat{p}_k(s)\ge \tau_k]$.

For supervised encoders (and QLoRA models that output probabilities), we use (i) a fixed global threshold $\tau=0.5$ or (ii) label-wise thresholds $\{\tau_k\}$ from a constrained grid search over $\tau \in \{0.00,0.01,\dots,1.00\}$. For each label $k$, we solve
\begin{equation}
\tau_k^*
=
\arg\max_{\tau \in \{0.00,0.01,\dots,1.00\}}
\mathrm{Recall}_k(\tau)
\quad
\text{s.t.}
\quad
\mathrm{Precision}_k(\tau)\ge 0.40.
\label{eq:threshold_optimization}
\end{equation}
This turns threshold selection into a constrained decision rule: under severe imbalance, we preserve a minimum precision floor while maximizing recall, which is often the limiting factor for Macro-$F_1$ on sparse labels. For hierarchical models, thresholds are tuned in stage-aware order (\emph{Presence} first, then HO, then values). Prompted LLMs output discrete label sets and do not require threshold calibration.

\subsection{Ensembling}
\label{sec:method_ensembles}

To test whether small diversity gains yield robust improvements, we evaluate simple ensembles over a pool of trained models:
\begin{itemize}
  \item \textbf{Hard voting}: average binary decisions.
  \item \textbf{Soft voting}: average predicted probabilities, then threshold (for families that output probabilities).
  \item \textbf{Weighted voting}: probability averages weighted by validation Macro-$F_1$ (probability-outputting families only).
\end{itemize}
We build ensembles via forward selection: start from the best single model on validation, then add a candidate only if it improves validation performance and the one-sided bootstrap lower 95\% bound for $\Delta F_1$ versus the current ensemble is $>0$ and at least $1\%$ in relative terms; ties go to the smaller ensemble. For prompted LLMs (discrete outputs), we use hard voting; for supervised encoders (and QLoRA), we also use soft and weighted voting.

\subsection{Evaluation metrics and statistical testing}
\label{sec:method_eval}

\paragraph{Primary metric.}
We report macro-averaged $F_1$ over the label set. We compute $F_1$ per label across all sentences, then average across labels. Macro-$F_1$ is appropriate here because the label distribution is strongly imbalanced and the study aims to evaluate gains on rare as well as frequent values; unlike micro-averaging, it prevents common labels from dominating the summary metric. For bipolar slices (Section~\ref{sec:method_problem}), we compute Macro-$F_1$ for each pole and average the two poles.

\paragraph{End-task evaluation.}
End-task (end-to-end) evaluation is the main metric: we compute Macro-$F_1$ on the full evaluation split using the \emph{final} system outputs for the target label space (e.g., the 19 values), where negative gate decisions force downstream predictions to zero (Eq.~\eqref{eq:cat_mask}). This captures both downstream quality and error propagation from upstream gating.

\paragraph{Uncertainty and significance.}
To assess robustness, we use nonparametric bootstrap resampling over sentences \citep{Efron1979}. We draw $B{=}2000$ samples with replacement, recompute Macro-$F_1$, and estimate (i) a one-sided 95\% lower bound for $\Delta F_1$ and (ii) a one-sided empirical $p$-value.

\paragraph{Per-label paired tests.}
For individual labels, we use McNemar's test on paired predictions to detect asymmetric error changes \citep{McNemar1947}. We correct for multiple comparisons across labels using Benjamini--Hochberg FDR control \citep{BenjaminiHochberg1995} and report corrected significance where relevant.

\subsection{Reproducibility}
\label{sec:method_repro}

We log all configurations (preprocessing, model variants, thresholds, ensemble membership) and preserve prediction files for all runs. We also release the value-to-HO mapping and scripts for preprocessing and evaluation to enable exact replication on the benchmark splits \citep{Touche2024}. For LLM experiments, we release the prompts/templates and any adapter weights (where redistribution is permitted), along with deterministic decoding settings and post-processing scripts.

\begin{figure}[htbp]
    \centering
    \includegraphics[width=\linewidth]{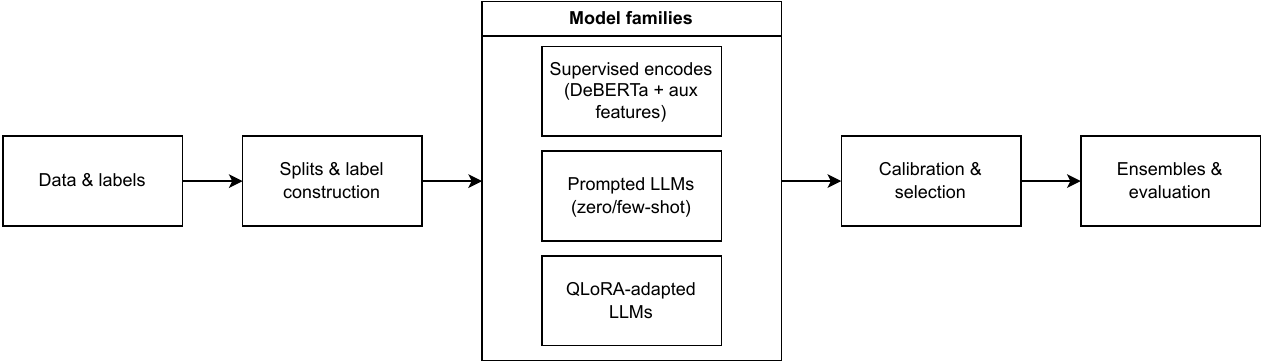}
    \caption{Overview of the experimental pipeline.}
    \label{fig:experimental_pipeline}
\end{figure}

Figure~\ref{fig:experimental_pipeline} summarizes the pipeline. We start from the English ValueEval'24/ValuesML release, construct basic-value, HO, and Presence labels, and use the official train/validation/test splits. We then train or evaluate three model families under an 8\,GB GPU budget (supervised encoders, prompted instruction-tuned LLMs, QLoRA-adapted LLMs), calibrate thresholds on validation, and select champion models. Finally, we form small ensembles and evaluate all systems with macro-$F_1$ and paired significance tests (bootstrap and McNemar) on both HO and basic-value slices.

\section{Results}
\label{sec:results}

We organize results around the research questions from Section~\ref{sec:introduction}: we first address HO category learnability (RQ1) and compute-frugal upgrades (RQ4), then evaluate hierarchical mechanisms for downstream value prediction (RQ2--RQ3), and finally benchmark small instruction-tuned LLMs under the same budget (RQ5).

For completeness, Section S2 reports the full validation/test tables for all higher-order category experiments and ablations. Here we focus on the main trends.

Throughout this section, we use three criteria to assess whether a modeling choice is genuinely useful: (i) absolute end-task performance on the full held-out test distribution, (ii) robustness across HO slices rather than isolated wins, and (iii) paired statistical support from bootstrap and McNemar analyses. This distinction is important for hierarchical methods: a component can look strong on a restricted conditional subproblem while still failing to improve the final end-to-end task.

\subsection{Higher-order categories are learnable, but not equally so}
\label{sec:results_ho_learnable}

HO categories are learnable with compact supervised encoders, but difficulty varies by pair. The easiest pair is \emph{Growth vs.\ Self-Protection} (test Macro-$F_1 \approx 0.58$, Table S3). \emph{Self-Transcendence vs.\ Self-Enhancement} is moderate (best Macro-$F_1 \approx 0.51$, Table S9), while \emph{Openness vs.\ Conservation} is hardest (best Macro-$F_1 \approx 0.42$, Table S7). These patterns track label prevalence: \emph{Openness} is rare (about 8\% vs.\ about 20\% for \emph{Conservation}, Table~\ref{tab:prevalence_dims}), which likely drives both lower Macro-$F_1$ and strong pole asymmetry (\emph{Openness} $F_1 \ll$ \emph{Conservation} $F_1$ in Table~\ref{tab:summary_ho_test}).

Table~\ref{tab:summary_ho_test} summarizes these differences: \emph{Growth vs.\ Self-Protection} is most learnable (Macro-$F_1=0.58$), \emph{Self-Transcendence vs.\ Self-Enhancement} is moderate (best Macro-$F_1=0.51$), and \emph{Openness vs.\ Conservation} remains hardest (best Macro-$F_1 \approx 0.42$), with persistent asymmetry (\emph{Conservation} $>$ \emph{Openness}). The pattern suggests that “constraint/tradition” cues are captured more reliably than “novelty/autonomy” cues at sentence level. We report single-run results and interpret differences under \mbox{\textasciitilde1--2} Macro-$F_1$ points cautiously.

\begin{table}[!htbp]
\centering
\footnotesize
\caption{Test Macro-$F_1$ for HO category detection by bipolar pair, showing the baseline, best tuned result, and best setting.}
\label{tab:summary_ho_test}
\begin{tabularx}{\textwidth}{XcXc}
\hline
\textbf{HO pair} & \textbf{Baseline} & \textbf{Best model} & \textbf{Best tuned}\\
 & \textbf{$F_1$ Fixed/Tuned} & \textbf{(if $\neq$ baseline)} & \textbf{$F_1$/pole A/pole B}\\
\hline
Growth vs.\ Self-Protection & 0.58/0.58 & Baseline & \textbf{0.58}/0.54/0.62\\
Social Focus vs.\ Personal Focus & 0.41/0.43 & NER / WorryWords / LIWC15 & \textbf{0.57}/0.59/0.54\\
Openness to Change vs.\ Conservation & 0.42/0.38 & LIWC22 + Ling.\ Feat & \textbf{0.42}/0.34/0.50\\
Self-Transcendence vs.\ Self-Enhancement & 0.48/0.50 & WorryWords / MFD-20 & \textbf{0.51}/0.53/0.48\\
\hline
\end{tabularx}
\end{table}

A second asymmetry appears in per-class $F_1$: for harder pairs, one pole is much more learnable (e.g., \emph{Conservation} consistently outperforms \emph{Openness}, Table~S7). This suggests that models capture “normative constraint” language (rules, tradition, order) more reliably than implicit “novelty/autonomy” cues.

\subsection{Cheap knobs matter: threshold calibration is consistently helpful, except when it overfits}
\label{sec:results_thresholds}

Table~\ref{tab:summary_ho_test} shows the effect of fixed vs.\ tuned thresholds: tuning strongly helps \emph{Social Focus} vs.\ \emph{Personal Focus} (0.41 $\rightarrow$ 0.57) and yields a smaller but consistent gain for \emph{Self-Transcendence} vs.\ \emph{Self-Enhancement} (0.48 $\rightarrow$ 0.51), while it can overfit under severe imbalance for \emph{Openness} vs.\ \emph{Conservation} (0.417 $\rightarrow$ 0.38 for the baseline).

Threshold calibration is a low-cost lever with frequent gains. For \emph{Social Focus vs.\ Personal Focus}, tuned thresholds consistently improve Macro-$F_1$, reaching $\approx 0.57$ on test (e.g., NER/WorryWords/LIWC15, Table~S5). The fixed-0.5 baseline underperforms (0.41), suggesting sensitivity to calibration and/or distribution shift; tuned thresholds recover performance (Table~\ref{tab:summary_ho_test}). For \emph{Self-Transcendence vs.\ Self-Enhancement}, tuning improves from $\approx 0.48$ to $\approx 0.50$, with best variants at $\approx 0.51$ (Table~S9). In contrast, for \emph{Openness vs.\ Conservation}, tuned thresholds can reduce Macro-$F_1$ (Table~S7), consistent with per-label overfitting under severe imbalance (e.g., extreme thresholds for \emph{Openness}).

Overall, \emph{label-wise thresholds are usually worth it in compute-frugal regimes}, but highly imbalanced labels benefit from conservative tuning rules (or calibration methods that regularize thresholds toward a global prior).

\subsection{Lightweight auxiliary signals rarely move the needle, but can stabilize certain pairs}
\label{sec:results_features}

Most feature add-ons (lexica, topic features, short context) yield small or inconsistent improvements and rarely beat a well-tuned baseline. Still, a few trends emerge.

Table~\ref{tab:summary_ho_test} shows that auxiliary signals change top rankings for only a subset of pairs: \emph{Social Focus vs.\ Personal Focus} benefits from weakly supervised cues (NER and lexical resources reach $\approx 0.57$ with tuned thresholds), and \emph{Self-Transcendence vs.\ Self-Enhancement} shows modest gains (best $\approx 0.51$). In contrast, \emph{Growth vs.\ Self-Protection} is near its ceiling, and \emph{Openness vs.\ Conservation} remains difficult even with added features.

For \emph{Social vs.\ Personal}, weakly supervised cues help at test time (Table~S5): NER and affective/moral lexica (e.g., WorryWords, LIWC15) reach Macro-$F_1 \approx 0.57$. This likely reflects explicit references to social groups, institutions, or interpersonal relations that lexical and NER cues partially capture. For \emph{Self-Transcendence vs.\ Self-Enhancement}, lexical signals again help modestly (Macro-$F_1 \approx 0.51$, Table~S9), suggesting a small benefit from prosocial vs.\ status/power cues.

At the same time, ablations show that \emph{auxiliary signals can hurt sharply if misaligned or noisy}. This may reflect configuration issues (e.g., feature scaling or dimensional mismatch), but the takeaway is practical: \emph{cheap features are only cheap if they are robustly engineered}; otherwise they can dominate the classifier head and destabilize training.

\subsection{Presence gating boosts in-gate validation scores but does not robustly improve end-task performance}
\label{sec:results_presence_gate}

\emph{Presence} gating (Eq.~\eqref{eq:presence}) produces a large jump in in-gate validation Macro-$F_1$ when evaluation is restricted to sentences predicted to contain a value (Section~S3). For example, \emph{Growth vs.\ Self-Protection} rises from about $0.58$ to $\sim 0.77$, and \emph{Social vs.\ Personal} from about $0.54$ to $\sim 0.74$. This is expected: removing many easy negatives changes the operating point and reduces sparsity.

Table~\ref{tab:summary_presence_gate} shows the same pattern across all four HO pairs: restricting evaluation to gate-passing sentences increases validation Macro-$F_1$ by +0.14 to +0.16, but these gains do not translate into consistent improvements on the original test distribution.

\begin{table}[!htbp]
\centering
\footnotesize
\caption{Effect of hard Presence gating on HO category detection: validation direct vs.\ gated Macro-$F_1$ and best test direct vs.\ Presence-gated Macro-$F_1$.}
\label{tab:summary_presence_gate}
\begin{tabularx}{\textwidth}{Xcccc}
\hline
\textbf{HO pair} & \textbf{Val Macro-$F_1$} & \textbf{Test Cascading Macro-$F_1$}\\
 & \textbf{Direct / Presence gate} & \textbf{Direct / Presence-gated}\\
\hline
Growth vs.\ Self-Protection & 0.58 / 0.77 & \textbf{0.58} / 0.58 \\
Social Focus vs.\ Personal Focus & 0.54 / 0.74 & \textbf{0.57} / 0.56 \\
Openness to Change vs.\ Conservation & 0.42 / 0.58 & 0.42 / \textbf{0.43} \\
Self-Transcendence vs.\ Self-Enhancement & 0.45 / 0.59 & \textbf{0.51} / 0.50 \\
\hline
\end{tabularx}
\end{table}

However, on the original test distribution, \emph{Presence gating does not yield consistent improvements}. The best gated Macro-$F_1$ matches the direct baseline for \emph{Growth/Self-Protection} (0.58), slightly underperforms for Social/Personal (0.56 vs.\ 0.57) and \emph{Self-Transcendence/Self-Enhancement} (0.50 vs.\ 0.51), and yields only a marginal gain for \emph{Openness/Conservation} (0.43 vs.\ 0.42) (Table~\ref{tab:summary_presence_gate}).

Table~\ref{tab:presence_gate_sensitivity} adds a compact sensitivity check for the same hard-gating idea. Keeping the HO model fixed (\emph{Baseline}) and varying only the \emph{Presence} gate threshold from $\tau_g{=}0.50$ to $\tau_g{=}0.10$ within the same gate family (LIWC22+Ling.\ Feat.) does not overturn the conclusion: the gated system remains equal to or below the tuned \emph{Direct} champion in all four pairs, and class-wise HO threshold tuning does not provide a systematic rescue.

\begin{table}[!htbp]
\centering
\footnotesize
\caption{Compact sensitivity check for hard Presence gating on test Macro-$F_1$. `Direct best' is the tuned direct champion from Table~\ref{tab:summary_presence_gate}. Gated values report test Macro-$F_1$ as fixed/tuned HO thresholds for the same HO model (\emph{Baseline}) and the same gate family (LIWC22+Ling.\ Feat.), varying only the Presence gate threshold $\tau_g$. Full sweeps appear in Tables S11 and S13--S17.}
\label{tab:presence_gate_sensitivity}
\begin{tabularx}{\textwidth}{Xccc}
\hline
\textbf{HO pair} & \textbf{Direct best} & \textbf{$\tau_g{=}0.50$} & \textbf{$\tau_g{=}0.10$} \\
 &  & \textbf{fixed / tuned} & \textbf{fixed / tuned} \\
\hline
Growth vs.\ Self-Protection & 0.580 & 0.570 / 0.580 & 0.570 / 0.560 \\
Social Focus vs.\ Personal Focus & 0.570 & 0.560 / 0.560 & 0.550 / 0.550 \\
Openness to Change vs.\ Conservation & 0.420 & 0.412 / 0.384 & 0.416 / 0.414 \\
Self-Transcendence vs.\ Self-Enhancement & 0.510 & 0.490 / 0.490 & 0.490 / 0.480 \\
\hline
\end{tabularx}
\end{table}

The most plausible explanation is error compounding: \emph{Presence} false negatives suppress downstream positives (recall loss), while \emph{Presence} false positives admit sentences that still look negative to the HO classifier (precision loss). The net effect is that \emph{hard gating trades recall for precision in a way that is hard to tune globally}. The compact sweep in Table~\ref{tab:presence_gate_sensitivity} and the full appendix sweeps (Tables S11 and S13--S17) show the same pattern: changing the gate threshold or switching from fixed to tuned HO thresholds produces only minor fluctuations and does not reverse the direct-vs.-gated ranking. This motivates softer alternatives rather than a binary filter.

Paired tests corroborate these trends: in several HO slices the best tuned \emph{Direct} systems are significantly stronger than \emph{Presence}-based alternatives on test (Section~S7), indicating that \emph{Presence$\rightarrow$Category} cascades do not provide a robust out-of-sample advantage under hard gating.

\subsection{HO$\rightarrow$values hard gating does not translate into reliable gains on fine-grained detection}
\label{sec:results_ho_gate_values}

We next test whether HO predictions help fine-grained value detection by constraining the value space. Table~\ref{tab:summary_ho_values_hard_gating} summarizes the best test Macro-$F_1$ with \emph{hard} HO$\rightarrow$values gates across all HO categories (details in Section~S4). Overall, hard HO gating via a binary mask is not a “free win.” Even the best gated configurations remain modest and do not consistently outperform tuned \emph{Direct} baselines (Section~S7).

\begin{table}[!htbp]
\centering
\footnotesize
\setlength{\tabcolsep}{4pt}
\caption{Best test Macro-$F_1$ for \emph{hard} HO$\rightarrow$values gating by HO slice, with the best value-model and gate thresholds.}
\label{tab:summary_ho_values_hard_gating}
\begin{tabularx}{\textwidth}{lccX}
\hline
\textbf{HO gate} & \textbf{\#values} & \textbf{Best test Macro-$F_1$} & \textbf{Best configuration (value model / gate)} \\
\hline
Growth & 11 & 0.271 & Baseline@0.30 \;/\; Baseline@0.50 \\
Self-Protection & 10 & 0.306 & Baseline@0.15 \;/\; Baseline@0.50 \\
Social Focus & 10 & 0.326 & Lex - MJD@0.25 \;/\; TD - NMF@0.50 \\
Personal Focus & 9 & 0.287 & Lex - MJD@0.70 \;/\; NER@0.19 \\
Openness & 4 & 0.235 & NER + TD - LDA@0.50 \;/\; Lex - LIWC 22 + Ling Feat@0.50 \\
Conservation & 7 & 0.297 & Lex - Schwartz@0.50 \;/\; Lex - LIWC 22 + Ling Feat@0.10 \\
Self-Transcendence & 6 & 0.307 & Baseline@0.25 \;/\; Lex - MFD-20@0.50 \\
Self-Enhancement & 5 & 0.320 & Lex - Schwartz@0.55 \;/\; Lex - WorryWords@0.23 \\
\hline
\end{tabularx}
\end{table}

Under the \emph{Growth} gate, the best test Macro-$F_1$ for \emph{Growth} values is about $0.271$; under \emph{Self-Protection} it is about $0.306$. Section~S7 compares the best HO-gated systems against tuned \emph{Direct} baselines within each HO slice using paired bootstrap tests. The pattern is consistent: hard HO gating does not yield reliable improvements and is sometimes significantly worse, consistent with recall suppression when parent categories are missed. This brittleness is not tied to a single threshold choice: Table~\ref{tab:summary_ho_values_hard_gating} already reports the best-performing gated configurations over a range of gate thresholds (from 0.10 to 0.50), yet even these threshold-selected variants do not produce consistent downstream gains. In a sentence-level setting—where signals are short, implicit, and often multi-valued—this loss is hard to recover.

In short, HO structure is informative, but \emph{forcing} predictions to respect the hierarchy via a binary mask is often too rigid for noisy sentence-level supervision.

\subsection{Where small instruction-tuned LLMs fit under the same budget}
\label{sec:results_llms_main}

Section~S4 benchmarks instruction-tuned $\le$10B LLMs (prompted and QLoRA-adapted) on the same HO-restricted slices. Table~\ref{tab:llm_budget_summary} reports the best test Macro-$F_1$ for Gemma-2-9B-it under (i) prompting only, (ii) prompting plus a lightweight SBERT gate, and (iii) QLoRA adaptation, and indicates which upgrades are supported by bootstrap tests (Section~S7). Overall, even the best LLM variants remain well below the supervised DeBERTa-based champions on test (e.g., \emph{Growth}: $0.201$ vs.\ $0.303$; \emph{Self-Protection}: $0.254$ vs.\ $0.342$; \emph{Social Focus}: $0.267$ vs.\ $0.345$; \emph{Personal Focus}: $0.198$ vs.\ $0.317$; see Tables~\ref{tab:llm_budget_summary}, \ref{tab:ens_ho_summary}).

\begin{table*}[!t]
\centering
\footnotesize
\setlength{\tabcolsep}{4pt}
\caption{Gemma 2 9B IT results by HO slice. Sig.: FS = few-shot vs.\ zero-shot, G = SBERT gate, Q = QLoRA, E = LLM ensemble, X = Transformer+LLM ensemble; ``+'' significant, ``0'' not significant or negative, ``--'' not tested.}
\label{tab:llm_budget_summary}
\begin{tabularx}{\textwidth}{lccccX}
\hline
\textbf{HO slice} & \textbf{Prompted} & \textbf{+gate} & \textbf{QLoRA} & \textbf{Ensemble} & \textbf{Sig.} \\
\hline
Growth & 0.191 & 0.194 & 0.132 & \textbf{0.201} & FS+, G+, Q0, E+, X-- \\
Self-Protection & 0.220 & 0.212 & \textbf{0.254} & 0.254 & FS+, G0, Q+, E--, X+ \\
Social Focus & 0.230 & 0.234 & 0.206 & \textbf{0.267} & FS+, G+, Q0, E+, X-- \\
Personal Focus & \textbf{0.198} & 0.195 & 0.184 & 0.198 & FS+, G0, Q0, E--, X+ \\
Openness & 0.093 & \textbf{0.113} & 0.123 & 0.123 & FS+, G+, Q0, E--, X0 \\
Conservation & 0.230 & 0.227 & \textbf{0.258} & 0.258 & FS+, G+, Q+, E--, X+ \\
Self-Transcendence & 0.213 & \textbf{0.221} & 0.124 & 0.221 & FS+, G+, Q0, E0, X-- \\
Self-Enhancement & \textbf{0.221} & 0.222 & 0.227 & 0.227 & FS+, G--, Q--, E--, X-- \\
\hline
\end{tabularx}
\end{table*}

Bootstrap comparisons in Section~S7 show that (i) few-shot prompting significantly improves over zero-shot in multiple slices (FS+; e.g., \emph{Growth, Social Focus, Self-Protection, Personal Focus}), and (ii) adding an SBERT-style gate can yield additional gains in some settings (G+; e.g., \emph{Growth} and \emph{Social Focus}), but not reliably in others (e.g., \emph{Personal Focus}, \emph{Self-Protection}). QLoRA adaptation is \emph{mixed}: it improves in \emph{Self-Protection} and \emph{Conservation}, but degrades in \emph{Growth} and \emph{Social Focus}, suggesting sensitivity to sparsity and slice-specific shifts. Where reported, within-family LLM ensembling can also help (E+; \emph{Growth}, \emph{Social Focus}).

LLMs can still be useful as a \emph{diversity source} in cross-family ensembles. For \emph{Self-Protection} and \emph{Personal Focus}, combining the best transformer with the best LLM yields a further significant improvement (X+), indicating complementary error patterns despite weaker standalone performance.

\subsection{Simple ensembling is the most reliable compute-frugal gain}
\label{sec:results_ensembles}

Across HO slices, the strongest and most repeatable improvements come from \emph{small, low-cost ensembles} rather than hard hierarchical masks. Table~\ref{tab:ens_ho_summary} summarizes test Macro-$F_1$ for the transformer champions ensemble (soft voting) and the corresponding bootstrap outcomes (Sections S6 and S7). Overall, ensembling yields consistent point gains, with the clearest improvements in \emph{Growth, Self-Protection, and Personal Focus}.

\begin{table}[!htbp]
\centering
\caption{Test Macro-$F_1$ of compute-frugal ensembles on HO slices. Sig.: ``+'' significant, ``0'' not significant or negative, ``--'' not tested.}
\label{tab:ens_ho_summary}
\begin{tabularx}{\textwidth}{l c c c c c}
\hline
\textbf{HO slice} & \textbf{Single} & \textbf{Tr ens.; Sig.} & \textbf{Tr+LLM; Sig.} \\
\hline
Growth & 0.286 & \textbf{0.303; +} & --; -- \\
Self-Protection & 0.321 & 0.342; + & \textbf{0.353; +} \\
Social Focus & \textbf{0.330} & 0.345; 0 & --; -- \\
Personal Focus & 0.305 & 0.317; + & \textbf{0.326; +} \\
Openness to Change & \textbf{0.240} & 0.241; 0 & 0.249; 0 \\
Conservation & 0.301 & 0.313; 0 & \textbf{0.334; +} \\
Self-Transcendence & \textbf{0.302} & 0.319; 0 & --; -- \\
Self-Enhancement & \textbf{0.338} & 0.337; -- & 0.349; 0 \\
\hline
\end{tabularx}
\end{table}

Section~S7 confirms that these ensemble gains are not just noise. For example, in \emph{Growth}, moving from the tuned \emph{Direct} baseline to the transformer ensemble increases Macro-$F_1$ from $0.286$ to $0.303$ and is significant. \emph{Self-Protection} and \emph{Personal Focus} show similar significant lifts over the best single model in the paired comparison.

Not all slices benefit equally: for \emph{Social Focus}, the ensemble improves the point estimate but does not meet the one-sided bootstrap criterion. Similar non-significant (but positive) gains appear in \emph{Openness, Conservation, and Self-Transcendence} (Table~\ref{tab:ens_ho_summary}).

\subsection{Which improvements are statistically robust under fixed compute}
\label{sec:results_significance_summary}

Section~S7 evaluates paired differences with a one-sided bootstrap test on $\Delta$Macro-$F_1$ and per-label McNemar tests with Benjamini--Hochberg correction. Table~\ref{tab:fixedcompute-robust-summary} condenses the main fixed-compute robustness results across all slices. Three consistent patterns emerge.

\begin{table}[!htbp]
\centering
\footnotesize
\caption{Fixed-compute bootstrap results across HO slices. Entries report lower 95\% bounds on $\Delta$Macro-$F_1$ and significance; \texttt{+} significant, \texttt{0} not significant or negative, \texttt{--} not tested.}
\label{tab:fixedcompute-robust-summary}
\begin{tabularx}{\textwidth}{l c c c c}
\hline
\textbf{HO Slice} & \textbf{Direct vs gated; Sig.} & \textbf{Ensemble; Sig.} & \textbf{Hybrid; Sig.} \\
\hline
Growth (HO) & 0.003; \texttt{+} & 0.002; \texttt{+} & \texttt{--} \\
Self-Protection (HO) & 0.002; \texttt{+} & 0.006; \texttt{+} & 0.007; \texttt{+} \\
Social Focus (HO) & -0.002; \texttt{0} & -0.001; \texttt{0} & \texttt{--} \\
Personal Focus (HO) & 0.011; \texttt{+} & 0.003; \texttt{+} & 0.005; \texttt{+} \\
Openness (HO) & -0.003; \texttt{0} & -0.016; \texttt{0} & -0.010; \texttt{0} \\
Conservation (HO) & -0.009; \texttt{0} & -0.006; \texttt{0} & 0.011; \texttt{+} \\
Self-Transcendence (HO) & -0.016; \texttt{0} & -0.007; \texttt{0} & \texttt{--} \\
Self-Enhancement (HO) & 0.010; \texttt{+} & -0.014; \texttt{0} & -0.003; \texttt{0} \\
\hline
\end{tabularx}
\end{table}

First, \emph{threshold tuning is a statistically reliable improvement}: in every slice with reported tests, tuned thresholds significantly outperform fixed $\tau{=}0.5$ for \emph{Direct} models (see ``Thr.\ tuning'' in Table~\ref{tab:fixedcompute-robust-summary}). McNemar analyses show these gains concentrate in subsets of labels (e.g., Universalism/Benevolence in \emph{Self-Transcendence}; Security/Conformity/Tradition in \emph{Conservation}; several sparse HO labels), rather than uniformly.

Second, \emph{hard hierarchical gating is not a reliable downstream win}. For HO slices, the tuned \emph{Direct} champion significantly outperforms the HO-gated champion in three of four cases (\emph{Growth, Self-Protection, Personal Focus}); \emph{Social Focus} shows no significant difference. For HO categories, none of the reported gated champions beats the \emph{Direct} champion (\emph{Openness, Conservation, Self-Transcendence}), consistent with error compounding under hard masks.

Third, \emph{ensembles provide the most consistent significant gains beyond threshold tuning, mainly in HO slices}. Transformer soft-voting ensembles yield significant improvements in \emph{Growth, Self-Protection, and Personal Focus}, while \emph{Social Focus} (HO) and all reported HO categories show non-significant uplift. Hybrid ensembles can yield additional gains in some cases (\emph{Self-Protection, Personal Focus, Conservation}), but not universally (e.g., \emph{Openness}).

Taken together, these results give a coherent interpretation of the study design and findings. \textbf{H1} is supported in qualified form: HO categories are learnable, but their usefulness depends strongly on prevalence and lexical concentration. \textbf{H2} is supported more clearly: hard hierarchical routing improves structural consistency and can inflate conditional scores, yet it does not deliver robust end-task gains because recall losses accumulate across stages. \textbf{H3} receives the strongest empirical support: under fixed compute, threshold tuning and small ensembles are the only interventions that repeatedly yield statistically supported improvements. The significance of the paper therefore lies not in proposing a new hierarchy method, but in establishing a more precise benchmark-level conclusion: HO structure is most useful as a descriptive or auxiliary inductive bias, whereas the practically reliable improvements come from calibration and lightweight ensembling.

\section{Discussion}
\label{sec:discussion}

The results do more than rank systems: they clarify \emph{why} some strategies help and others fail under sentence-level sparsity, overlap, and fixed compute. In particular, the experiments separate the representational value of HO abstractions from the decision-theoretic cost of enforcing them as hard gates. This distinction is central to the paper's novelty framing, because the main contribution is not a single architecture, but a careful empirical analysis of how HO structure should be used in practice under a bounded compute budget.

Taken together, the experiments point to five practical outcomes.

\paragraph{(1) HO abstractions are learnable, but they are not uniformly reliable.}
Pairs with higher prevalence and stronger lexical regularities (e.g., \emph{Growth/Self-Protection}) are easier; rare or diffuse categories (Openness) remain difficult even with tuning. For example, \emph{Growth/Self-Protection} reaches Macro-$F_1$ $\approx 0.58$, while \emph{Openness/Conservation} peaks around $\approx 0.42$ with persistent pole asymmetry (\emph{Conservation} $>$ \emph{Openness}).

\paragraph{(2) Calibration and small ensembles are safer bets than hard hierarchies.}
Threshold tuning yields small but frequent gains, and forward-selected soft-voting ensembles provide the most consistent significant improvements (Section~S7), while most feature add-ons are marginal or unstable. In HO detection, tuning ranges from modest gains (e.g., 0.48 $\rightarrow$ 0.51) to large calibration-sensitive jumps (e.g., Social/Personal 0.41 $\rightarrow$ 0.57); ensembles yield smaller but more reliable lifts (e.g., \emph{Growth} 0.286 $\rightarrow$ 0.303).

\paragraph{(3) \emph{Presence} gating and HO gating improve \emph{conditional} performance but not \emph{end-task} performance.}
Large validation gains under gating are largely an artifact of evaluating a simplified subproblem (value-present sentences) and do not carry over on the full test distribution. \emph{Presence} gating inflates in-gate validation Macro-$F_1$ by roughly +0.14 to +0.16, but test improvements are negligible or negative in most slices. A unifying explanation is error compounding: parent false negatives suppress child recall, parent false positives admit hard negatives, and imbalance makes threshold search volatile for rare poles (e.g., \emph{Openness}).

\paragraph{(4) Hard gates do not yield reliable end-to-end gains.}
We directly test \emph{hard} hierarchical mechanisms (\emph{Presence} gating and HO$\rightarrow$values masking). Across HO pairs and value slices, these hard constraints do not yield reliable end-to-end gains and are sometimes significantly worse than tuned \emph{Direct} models, despite strong conditional scores. This is consistent with error propagation: uncertain parent decisions become binary filters that suppress true positives and hurt recall for sparse labels. While we do not evaluate \emph{soft} hierarchical conditioning here, the results motivate treating HO structure as an uncertainty-preserving inductive bias (e.g., probabilistic conditioning or auxiliary HO objectives) rather than a strict routing rule, and exploring broader context to reduce ambiguity.

\paragraph{(5) Small LLMs are not competitive alone, but can add useful diversity}
Under the same budget, prompted and QLoRA-adapted $\le$10B LLMs underperform supervised encoders in absolute Macro-$F_1$, although few-shot prompting helps. Their main practical benefit is as complementary signals in cross-family ensembles for some slices (Section~S7). For instance, Gemma-2-9B-it remains below transformer champions (e.g., \emph{Growth} $\approx0.20$ vs.\ $\approx0.30$), but can still improve cross-family ensembles in selected slices.

Beyond this benchmark, our results show that \emph{how} domain knowledge is injected matters as much as \emph{which} knowledge is used. Enforcing the HO taxonomy as a hard constraint may raise precision in a restricted space but can reduce end-task recall through error propagation. From a system-design standpoint, psychologically grounded taxonomies such as Schwartz values are best leveraged as \emph{regularizers and priors} rather than strict filters when predictions are noisy or labels overlap. Overall, enforcing hierarchy via hard gating is brittle in sentence-level, imbalanced, multi-valued settings.

\subsection{Limitations and threats to validity}
\label{sec:discussion_limitations}

Our findings have several limitations. First, we report single-run results, so differences of 1--2 Macro-$F_1$ points may be unstable. Second, the sentence-level setting limits signal; hierarchy may help more with broader discourse context. Annotation noise and multi-label overlap can also conflict with strict parent--child constraints. Third, calibration can overfit under severe imbalance, especially for rare categories (e.g., \emph{Openness}). Finally, this is a benchmark-driven study: the conclusions are tied to ValueEval'24 / ValuesML and a compute-frugal regime, so external validity across domains, languages, annotation schemes, or larger-model settings remains to be established in future work.

\subsection{Answers to the research questions}
\label{sec:discussion_rqs}

\textbf{RQ1 (Are HO values learnable from single sentences?).} Yes---HO categories are learnable with compact supervised encoders, but learnability varies widely across pairs; rare/diffuse categories (e.g., \emph{Openness}) remain challenging under fixed compute (Section~\ref{sec:results_ho_learnable}).

\textbf{RQ2 (Do HO gates help downstream basic-value prediction?).} Under \emph{hard} masking (Category$\rightarrow$Values), HO gating does not reliably improve out-of-sample Macro-$F_1$ and can be significantly worse than tuned \emph{Direct} models (Section~S7), consistent with error compounding. This negative result is specific to \emph{hard} masking and does not rule out gains from \emph{soft} HO integration, which we do not evaluate here.

\textbf{RQ3 (Does Presence$\rightarrow$Category outperform Category-only?).} With \emph{hard} \emph{Presence}-gated cascades, \emph{Presence} improves \emph{conditional} performance but the full pipeline does not consistently beat tuned \emph{Direct} baselines on the test distribution (Section~\ref{sec:results_presence_gate}). Gains are not robust across slices (Section~S7). This negative result is specific to binary gating and does not preclude improvements from learned or soft gates or end-to-end training.

\textbf{RQ4 (Which low-cost knobs move the needle?).} Threshold calibration is the most consistently significant improvement, and simple soft-voting ensembles provide additional gains in several HO slices (Sections~\ref{sec:results_ensembles}--\ref{sec:results_significance_summary}). Lexica/topic/context features are unstable: they can help specific slices but are not the main drivers under fixed compute.

\textbf{RQ5 (Where do small LLMs fit?).} Prompted $\le$10B LLMs benefit from few-shot prompting and sometimes from lightweight semantic gates, but they lag behind supervised DeBERTa-based models under the same budget (Section~\ref{sec:results_llms_main}). Their practical value is mainly as complementary signals in cross-family ensembles, which can yield significant improvements in some slices (Section~S7).

\section{Conclusions and future work}
\label{sec:conclusions-future-work}

This paper presented a compute-bounded empirical study of whether \emph{higher-order} (HO) value abstractions improve sentence-level human value detection. The results show that HO categories are learnable, but their difficulty varies markedly with prevalence and lexical regularity. The most reliable improvements come from label-wise threshold calibration and small soft-voting ensembles, whereas hard hierarchical mechanisms such as \emph{Presence} gating and HO$\rightarrow$value masking do not robustly improve the end task despite looking stronger under conditional evaluation. Compact instruction-tuned LLMs also remain weaker than supervised encoders in absolute Macro-$F_1$, although they can still add useful diversity in some cross-family ensembles. Overall, the benchmark-level conclusion is that HO structure is useful as an inductive bias, but too brittle when enforced as a hard routing rule for sparse, noisy, multi-label sentence classification.

Future work should therefore focus on hierarchy-aware methods that preserve uncertainty instead of discarding it. Promising directions include joint hierarchical learning, soft HO priors that condition value predictions on HO probabilities rather than binary masks, and stronger calibration methods for rare labels. Because sentence-level inputs often underrepresent value cues, future studies should also vary the amount of context available to the model and test these approaches across additional domains and annotation schemes. In other words, external validity beyond this benchmark-driven setting remains an open question for future work.

\printcredits

\section{Declaration of generative AI and AI-assisted technologies in the manuscript preparation process}
\label{sec:declaration-generative-ai}

During the preparation of this work, the authors used ChatGPT from OpenAI in order to improve the readability and language of the manuscript. After using this tool/service, the authors reviewed and edited the content as needed and take full responsibility for the content of the published article.

\section{Data availability}
\label{sec:data-availability}

This study uses the English, machine-translated ValueEval'24/ValuesML release \citep{Mirzakhmedova2024,ValueEval24Zenodo}.
The dataset is distributed under a Data Usage Agreement that allows research use but does not permit redistribution of the texts.
Accordingly, we cannot share the sentences or any derivative file that contains original textual content.
Researchers with appropriate access can obtain the official train/validation/test splits by registering and downloading the release from Zenodo \citep{ValueEval24Zenodo}.

To maximize reproducibility without violating the license, we release the full experimental pipeline and all non-text artifacts needed to reproduce our results.
Specifically, we release (i) the codebase for preprocessing, model training, evaluation, threshold calibration, and ensembling, with configuration files for architectures and hyperparameters; (ii) trained model artifacts, including fine-tuned DeBERTa checkpoints (direct predictors, gating components, feature-augmented variants) and QLoRA adapter weights for Gemma~2~9B; and (iii) inference outputs for every system (validation and test), including predicted probabilities, binarized decisions from selected thresholds, and the thresholds themselves (global and per-label where applicable).
These resources are provided via GitHub\footnote{\url{https://github.com/VictorMYeste/human-value-detection}} and Hugging Face\footnote{\url{https://huggingface.co/papers/2601.14172}}.

All released files are keyed only by the dataset's official identifiers (e.g., \texttt{Text-ID} and \texttt{Sentence-ID}) and contain no original sentences.
This allows any researcher with licensed access to ValueEval'24 to reproduce our tables and figures and build further analyses on the same splits.

\appendix
\setcounter{table}{0}
\renewcommand{\thetable}{A\arabic{table}}
\renewcommand{\theHtable}{A\arabic{table}}

\section{Value-to-HO mapping}
\label{app:ho-mapping}

Table~\ref{tab:ho-mapping} reports the fixed mapping used in this paper between Schwartz's 19 refined basic values and the eight HO categories.
Note that in the refined theory some basic values contribute to more than one HO category (e.g., \emph{Hedonism}, \emph{Achievement}, \emph{Face}, \emph{Humility}), so overlaps across rows are expected.

\begin{table}[!htbp]
\centering
\small
\setlength{\tabcolsep}{6pt}
\renewcommand{\arraystretch}{1.15}
\caption{Mapping from the 19 basic values to the eight HO categories; overlaps follow Schwartz's refined theory.}
\label{tab:ho-mapping}
\begin{tabularx}{\textwidth}{>{\raggedright\arraybackslash}p{0.28\textwidth} X}
\hline
\textbf{HO category} & \textbf{Basic values included} \\
\hline
Growth &
Humility; Benevolence: caring; Benevolence: dependability; Universalism: concern; Universalism: nature; Universalism: tolerance; Self-direction: thought; Self-direction: action; Stimulation; Hedonism; Achievement. \\

Self-Protection &
Achievement; Power: dominance; Power: resources; Face; Security: personal; Security: societal; Tradition; Conformity: rules; Conformity: interpersonal; Humility. \\

Social Focus &
Security: societal; Tradition; Conformity: rules; Conformity: interpersonal; Humility; Benevolence: caring; Benevolence: dependability; Universalism: concern; Universalism: nature; Universalism: tolerance. \\

Personal Focus &
Self-direction: thought; Self-direction: action; Stimulation; Hedonism; Achievement; Power: dominance; Power: resources; Face; Security: personal. \\

Openness to Change &
Self-direction: thought; Self-direction: action; Stimulation; Hedonism. \\

Self-Enhancement &
Hedonism; Achievement; Power: dominance; Power: resources; Face. \\

Conservation &
Face; Security: personal; Security: societal; Tradition; Conformity: rules; Conformity: interpersonal; Humility. \\

Self-Transcendence &
Humility; Benevolence: caring; Benevolence: dependability; Universalism: concern; Universalism: nature; Universalism: tolerance. \\
\hline
\end{tabularx}
\end{table}

\section{Label prevalence across data splits}
\label{app:prevalence}

This appendix reports the prevalence of each label in the train/validation/test splits. Prevalence is computed at the \emph{sentence level} as the percentage of sentences annotated with a given label. The \emph{Presence} row corresponds to the percentage of sentences with at least one label.

\begin{table}[!htbp]
\centering
\caption{Sentence-level prevalence (\%) of the fine-grained values by split.}
\label{tab:prevalence_fine}
\begin{tabular}{lccc}
\hline
\textbf{Label} & \textbf{Train} & \textbf{Validation} & \textbf{Test} \\
\hline
Self-direction: thought        & 1.29 & 1.15 & 1.17 \\
Self-direction: action         & 3.61 & 3.26 & 3.51 \\
Stimulation                    & 2.62 & 2.82 & 2.55 \\
Hedonism                       & 0.86 & 0.67 & 0.86 \\
Achievement                    & 6.42 & 6.37 & 6.25 \\
Power: dominance               & 4.63 & 4.40 & 4.33 \\
Power: resources               & 5.00 & 4.86 & 5.53 \\
Face                           & 1.81 & 1.90 & 1.83 \\
Security: personal             & 2.03 & 1.87 & 2.42 \\
Security: societal             & 8.95 & 8.46 & 7.90 \\
Tradition                      & 1.20 & 1.84 & 1.35 \\
Conformity: rules              & 6.10 & 6.41 & 6.25 \\
Conformity: interpersonal      & 1.35 & 1.37 & 1.34 \\
Humility                       & 0.24 & 0.29 & 0.21 \\
Benevolence: caring            & 2.29 & 2.29 & 2.22 \\
Benevolence: dependability     & 1.94 & 1.93 & 1.98 \\
Universalism: concern          & 4.97 & 4.50 & 5.04 \\
Universalism: nature           & 2.05 & 2.57 & 2.01 \\
Universalism: tolerance        & 1.07 & 0.81 & 1.17 \\
\hline
Presence                       & 51.53 & 50.99 & 50.81 \\
\hline
\end{tabular}
\end{table}

\begin{table}[!htbp]
\centering
\caption{Sentence-level prevalence (\%) of the HO categories by split.}
\label{tab:prevalence_dims}
\begin{tabular}{lccc}
\hline
\textbf{Dimension} & \textbf{Train} & \textbf{Validation} & \textbf{Test} \\
\hline
Growth            & 25.56 & 24.68 & 25.16 \\
Self-Protection   & 35.20 & 35.41 & 34.82 \\
Social Focus      & 28.19 & 28.48 & 27.50 \\
Personal Focus    & 26.59 & 25.54 & 26.72 \\
Openness to Change & 8.20  & 7.72  & 7.90  \\
Conservation      & 20.90 & 21.40 & 20.34 \\
Self-Transcendence & 12.16 & 11.93 & 12.27 \\
Self-Enhancement  & 18.03 & 17.62 & 18.24 \\
\hline
\end{tabular}
\end{table}

\section{Prompt templates for reported LLM experiments}
\label{app:llm-prompts}

This appendix records the prompt format used for the prompted-LLM results reported in Section~\ref{sec:results_llms_main}. The prompt family was inherited from the controlled prompt comparison in \citet{Yeste2026}, where the definition-style template outperformed simpler direct, QA-style, and hidden-CoT variants under the same compute-frugal regime. We therefore document the best-performing template used in this paper rather than re-running a full prompt-engineering sweep.

\paragraph{System prompt.}
\begin{quote}\small
\texttt{You are a moral-psychology assistant. Using the refined basic values taxonomy (Schwartz 1992; Schwartz et al.\ 2012), answer the user's labeling requests exactly as instructed.}
\end{quote}

\paragraph{Zero-shot user prompt (definition style).}
\begin{quote}\small\ttfamily
\#\#\# Value definitions\\
- <value\_1>: <one-line definition>\\
\ldots\\
- <value\_m>: <one-line definition>\\[0.3em]
\#\#\# Task\\
Identify which of the above values the SENTENCE relates to. Return only a JSON array of the matching value names.\\[0.3em]
SENTENCE: <sentence>
\end{quote}

The definition block contains one-line descriptions of the candidate labels derived from \citet{Schwartz2012overview}. For slice-restricted runs, the candidate list is limited to the values relevant to that slice. For example, one \emph{Self-Enhancement} run used the label subset \texttt{[Hedonism, Achievement, Power: dominance, Power: resources, Face]}.

\paragraph{Few-shot wrapper.}
\begin{quote}\small\ttfamily
SENTENCE: <example sentence>\\
OUTPUT: <example JSON array>\\
---\\
\end{quote}

For few-shot prompting, $k$ exemplars with the above schema are prepended before the final zero-shot query, with $k\in\{1,2,4,8,16,20\}$. When $k=20$, the exemplar pool includes at least one null example whose output is \texttt{[]}.

\paragraph{Decoding and parsing.}
All prompted runs use greedy decoding with \texttt{max\_new\_tokens}=200. Outputs are parsed as JSON arrays and mapped to labels by exact string matching. Invalid JSON outputs or out-of-vocabulary labels are treated as empty predictions. For HO evaluation, predicted basic values are mapped to HO labels via Eq.~\eqref{eq:ho_or}.

\bibliographystyle{cas-model2-names}
\bibliography{references}

@article{Schwartz2012,
  title={Refining the theory of basic individual values.},
  author={Schwartz, Shalom H and Cieciuch, Jan and Vecchione, Michele and Davidov, Eldad and Fischer, Ronald and Beierlein, Constanze and Ramos, Alice and Verkasalo, Markku and L{\"o}nnqvist, Jan-Erik and Demirutku, Kursad and others},
  journal={Journal of personality and social psychology},
  volume={103},
  number={4},
  pages={663},
  year={2012},
  publisher={American Psychological Association}
}

@article{Schwartz2012overview,
  author    = {Schwartz, Shalom H.},
  title     = {An Overview of the Schwartz Theory of Basic Values},
  journal = {Online Readings in Psychology and Culture,},
  volume    = {2},
  number    = {1},
  year      = {2012},
  doi       = {10.9707/2307-0919.1116}
}

@misc{Touche2024,
  author       = {Touch{\'e}},
  howpublished = {Web page},
  year         = {2024},
  url          = {https://touche.webis.de/semeval24/touche24-web/index.html}
}

@misc{ValueEval24Zenodo,
  author  = {{The ValuesML Team}},
  title   = {Touch{\'e}24{-}ValueEval},
  year    = {2024},
  month   = {8},
  version = {2024-08-09},
  publisher = {Zenodo},
  doi     = {10.5281/zenodo.13283288},
}

@Inbook{Efron1979,
  author="Efron, Bradley",
  title="Bootstrap Methods: Another Look at the Jackknife",
  bookTitle="Breakthroughs in Statistics: Methodology and Distribution",
  year="1992",
  publisher="Springer New York",
  address="New York, NY",
  pages="569--593",
  chapter="20",
  isbn="978-1-4612-4380-9",
  doi="10.1007/978-1-4612-4380-9\_41",
}

@article{McNemar1947,
  title={Note on the Sampling Error of the Difference Between Correlated Proportions or Percentages},
  volume={12},
  DOI={10.1007/BF02295996},
  number={2},
  journal={Psychometrika},
  author={McNemar, Quinn},
  year={1947},
  pages={153–157}
}

@inproceedings{LoshchilovHutter2018,
  title={Decoupled Weight Decay Regularization},
  author={Ilya Loshchilov and Frank Hutter},
  booktitle={International Conference on Learning Representations},
  year={2019},
  url={https://openreview.net/forum?id=Bkg6RiCqY7},
}

@article{BenjaminiHochberg1995,
  author = {Benjamini, Yoav and Hochberg, Yosef},
  title = {Controlling the False Discovery Rate: A Practical and Powerful Approach to Multiple Testing},
  journal = {Journal of the Royal Statistical Society: Series B (Methodological)},
  volume = {57},
  number = {1},
  pages = {289-300},
  doi = {10.1111/j.2517-6161.1995.tb02031.x},
  year = {1995}
}

@misc{Yeste2026,
  title={Human Values in a Single Sentence: Moral Presence, Hierarchies, and Transformer Ensembles on the Schwartz Continuum}, 
  author={Víctor Yeste and Paolo Rosso},
  year={2026},
  eprint={2601.14172},
  archivePrefix={arXiv},
  primaryClass={cs.CL},
  url={https://arxiv.org/abs/2601.14172}, 
}

@inproceedings{He2021,
  title={DeBERTa: Decoding-enhanced BERT with Disentangled Attention}, 
  author={Pengcheng He and Xiaodong Liu and Jianfeng Gao and Weizhu Chen},
  booktitle={International Conference on Learning Representations},
  year={2021},
  url={https://openreview.net/forum?id=XPZIaotutsD}, 
}

@misc{Hu2022,
  title={Lo{RA}: Low-Rank Adaptation of Large Language Models}, 
  author={Edward J. Hu and Yelong Shen and Phillip Wallis and Zeyuan Allen-Zhu and Yuanzhi Li and Shean Wang and Lu Wang and Weizhu Chen},
  booktitle={International Conference on Learning Representations},
  year={2022},
  url={https://openreview.net/forum?id=nZeVKeeFYf9}
}

@inproceedings{Dettmers2024,
  author = {Dettmers, Tim and Pagnoni, Artidoro and Holtzman, Ari and Zettlemoyer, Luke},
  booktitle = {Advances in Neural Information Processing Systems},
  editor = {A. Oh and T. Naumann and A. Globerson and K. Saenko and M. Hardt and S. Levine},
  pages = {10088--10115},
  publisher = {Curran Associates, Inc.},
  title = {QLoRA: Efficient Finetuning of Quantized LLMs},
  url = {https://proceedings.neurips.cc/paper\_files/paper/2023/file/1feb87871436031bdc0f2beaa62a049b-Paper-Conference.pdf},
  volume = {36},
  year = {2023}
}

@InProceedings{Dietterich2000,
  author="Dietterich, Thomas G.",
  title="Ensemble Methods in Machine Learning",
  booktitle="Multiple Classifier Systems",
  year="2000",
  publisher="Springer Berlin Heidelberg",
  address="Berlin, Heidelberg",
  pages="1--15",
  isbn="978-3-540-45014-6"
}

@book{Rokeach1973,
  author    = {Milton Rokeach},
  title     = {The Nature of Human Values},
  publisher = {Free Press},
  year      = {1973},
  address   = {New York}
}

@article{Schwartz1992,
  title = {Universals in the Content and Structure of Values: Theoretical Advances and Empirical Tests in 20 Countries},
  editor = {Mark P. Zanna},
  journal = {Advances in Experimental Social Psychology},
  publisher = {Academic Press},
  volume = {25},
  pages = {1-65},
  year = {1992},
  issn = {0065-2601},
  doi = {10.1016/S0065-2601(08)60281-6},
  author = {Shalom H. Schwartz}
}

@article{BardiSchwartz2003,
  author = {Anat Bardi and Shalom H. Schwartz},
  title ={Values and Behavior: Strength and Structure of Relations},
  journal = {Personality and Social Psychology Bulletin},
  volume = {29},
  number = {10},
  pages = {1207-1220},
  year = {2003},
  doi = {10.1177/0146167203254602},
  note ={PMID: 15189583}
}

@article{Lazer2009,
  title   = {Computational Social Science},
  author  = {Lazer, David and Pentland, Alex and Adamic, Lada and Aral, Sinan and Barab{\'a}si, Albert-L{\'a}szl{\'o} and Brewer, Devon and Christakis, Nicholas and Contractor, Noshir and Fowler, James and Gutmann, Myron and Jebara, Tony and King, Gary and Macy, Michael and Roy, Deb and Van Alstyne, Marshall},
  journal = {Science},
  volume  = {323},
  number  = {5915},
  pages   = {721--723},
  year    = {2009},
  doi     = {10.1126/science.1167742}
}

@article{HaidtJoseph2004,
  title   = {Intuitive Ethics: How Innately Prepared Intuitions Generate Culturally Variable Virtues},
  author  = {Haidt, Jonathan and Joseph, Craig},
  journal = {Daedalus},
  volume  = {133},
  number  = {4},
  pages   = {55--66},
  year    = {2004},
  doi     = {10.1162/0011526042365555}
}

@article{Graham2009,
  title   = {Liberals and Conservatives Rely on Different Sets of Moral Foundations},
  author  = {Graham, Jesse and Haidt, Jonathan and Nosek, Brian A.},
  journal = {Journal of Personality and Social Psychology},
  volume  = {96},
  number  = {5},
  pages   = {1029--1046},
  year    = {2009},
  doi     = {10.1037/a0015141}
}

@inproceedings{JohnsonGoldwasser2018,
    title = "Classification of Moral Foundations in Microblog Political Discourse",
    author = "Johnson, Kristen and Goldwasser, Dan",
    editor = "Gurevych, Iryna and Miyao, Yusuke",
    booktitle = "Proceedings of the 56th Annual Meeting of the Association for Computational Linguistics (Volume 1: Long Papers)",
    month = jul,
    year = "2018",
    address = "Melbourne, Australia",
    publisher = "Association for Computational Linguistics",
    doi = "10.18653/v1/P18-1067",
    pages = "720--730",
}

@inproceedings{Kiesel2023,
  title = "SemEval-2023 Task 4: ValueEval: Identification of Human Values Behind Arguments",
  author = "Kiesel, Johannes  and Alshomary, Milad  and Mirzakhmedova, Nailia  and Heinrich, Maximilian  and Handke, Nicolas  and Wachsmuth, Henning  and Stein, Benno",
  editor = {Ojha, Atul Kr.  and Do{\u{g}}ru{\"o}z, A. Seza  and Da San Martino, Giovanni  and Tayyar Madabushi, Harish  and Kumar, Ritesh  and Sartori, Elisa},
  booktitle = "Proceedings of the 17th International Workshop on Semantic Evaluation (SemEval-2023)",
  month = jul,
  year = "2023",
  address = "Toronto, Canada",
  publisher = "Association for Computational Linguistics",
  doi = "10.18653/v1/2023.semeval-1.313",
  pages = "2287--2303",
}

@inproceedings{TsoumakasKatakis2007,
  title     = {Multi-Label Classification: An Overview},
  author    = {Tsoumakas, Grigorios and Katakis, Ioannis},
  booktitle = {International Journal of Data Warehousing and Mining},
  volume    = {3},
  number    = {3},
  pages     = {1--13},
  year      = {2007},
  doi       = {10.4018/jdwm.2007070101}
}

@article{ZhangZhou2014,
  author={Zhang, Min-Ling and Zhou, Zhi-Hua},
  journal={IEEE Transactions on Knowledge and Data Engineering}, 
  title={A Review on Multi-Label Learning Algorithms}, 
  year={2014},
  volume={26},
  number={8},
  pages={1819-1837},
  doi={10.1109/TKDE.2013.39}
}

@Article{SillaFreitas2011,
  author="Silla, Carlos N. and Freitas, Alex A.",
  title="A survey of hierarchical classification across different application domains",
  journal="Data Mining and Knowledge Discovery",
  year="2011",
  month="Jan",
  day="01",
  volume="22",
  number="1",
  pages="31--72",
  issn="1573-756X",
  doi="10.1007/s10618-010-0175-9",
}

@InProceedings{Guo2017,
  title = 	 {On Calibration of Modern Neural Networks},
  author =       {Chuan Guo and Geoff Pleiss and Yu Sun and Kilian Q. Weinberger},
  booktitle = 	 {Proceedings of the 34th International Conference on Machine Learning},
  pages = 	 {1321--1330},
  year = 	 {2017},
  editor = 	 {Precup, Doina and Teh, Yee Whye},
  volume = 	 {70},
  series = 	 {Proceedings of Machine Learning Research},
  month = 	 {06--11 Aug},
  publisher =    {PMLR},
  url = 	 {https://proceedings.mlr.press/v70/guo17a.html},
}

@Article{Rokach2010,
  author="Rokach, Lior",
  title="Ensemble-based classifiers",
  journal="Artificial Intelligence Review",
  year="2010",
  month="Feb",
  day="01",
  volume="33",
  number="1",
  pages="1--39",
  issn="1573-7462",
  doi="10.1007/s10462-009-9124-7",
}

@inproceedings{Brown2020,
 author = {Brown, Tom and Mann, Benjamin and Ryder, Nick and Subbiah, Melanie and Kaplan, Jared D and Dhariwal, Prafulla and Neelakantan, Arvind and Shyam, Pranav and Sastry, Girish and Askell, Amanda and Agarwal, Sandhini and Herbert-Voss, Ariel and Krueger, Gretchen and Henighan, Tom and Child, Rewon and Ramesh, Aditya and Ziegler, Daniel and Wu, Jeffrey and Winter, Clemens and Hesse, Chris and Chen, Mark and Sigler, Eric and Litwin, Mateusz and Gray, Scott and Chess, Benjamin and Clark, Jack and Berner, Christopher and McCandlish, Sam and Radford, Alec and Sutskever, Ilya and Amodei, Dario},
 booktitle = {Advances in Neural Information Processing Systems},
 editor = {H. Larochelle and M. Ranzato and R. Hadsell and M.F. Balcan and H. Lin},
 pages = {1877--1901},
 publisher = {Curran Associates, Inc.},
 title = {Language Models are Few-Shot Learners},
 url = {https://proceedings.neurips.cc/paper\_files/paper/2020/file/1457c0d6bfcb4967418bfb8ac142f64a-Paper.pdf},
 volume = {33},
 year = {2020}
}

@inproceedings{Ouyang2022,
  author = {Ouyang, Long and Wu, Jeffrey and Jiang, Xu and Almeida, Diogo and Wainwright, Carroll and Mishkin, Pamela and Zhang, Chong and Agarwal, Sandhini and Slama, Katarina and Ray, Alex and Schulman, John and Hilton, Jacob and Kelton, Fraser and Miller, Luke and Simens, Maddie and Askell, Amanda and Welinder, Peter and Christiano, Paul F and Leike, Jan and Lowe, Ryan},
  booktitle = {Advances in Neural Information Processing Systems},
  editor = {S. Koyejo and S. Mohamed and A. Agarwal and D. Belgrave and K. Cho and A. Oh},
  pages = {27730--27744},
  publisher = {Curran Associates, Inc.},
  title = {Training language models to follow instructions with human feedback},
  url = {https://proceedings.neurips.cc/paper\_files/paper/2022/file/b1efde53be364a73914f58805a001731-Paper-Conference.pdf},
  volume = {35},
  year = {2022}
}

@article{Chung2024,
  author  = {Hyung Won Chung and Le Hou and Shayne Longpre and Barret Zoph and Yi Tay and William Fedus and Yunxuan Li and Xuezhi Wang and Mostafa Dehghani and Siddhartha Brahma and Albert Webson and Shixiang Shane Gu and Zhuyun Dai and Mirac Suzgun and Xinyun Chen and Aakanksha Chowdhery and Alex Castro-Ros and Marie Pellat and Kevin Robinson and Dasha Valter and Sharan Narang and Gaurav Mishra and Adams Yu and Vincent Zhao and Yanping Huang and Andrew Dai and Hongkun Yu and Slav Petrov and Ed H. Chi and Jeff Dean and Jacob Devlin and Adam Roberts and Denny Zhou and Quoc V. Le and Jason Wei},
  title   = {Scaling Instruction-Finetuned Language Models},
  journal = {Journal of Machine Learning Research},
  year    = {2024},
  volume  = {25},
  number  = {70},
  pages   = {1--53},
  url     = {http://jmlr.org/papers/v25/23-0870.html}
}

@article{Araque2020,
  title = {MoralStrength: Exploiting a moral lexicon and embedding similarity for moral foundations prediction},
  journal = {Knowledge-Based Systems},
  volume = {191},
  pages = {105184},
  year = {2020},
  issn = {0950-7051},
  doi = {10.1016/j.knosys.2019.105184},
  author = {Oscar Araque and Lorenzo Gatti and Kyriaki Kalimeri}
}

@article{GonzalezSantos2023,
  title = {Automatic assignment of moral foundations to movies by word embedding},
  journal = {Knowledge-Based Systems},
  volume = {270},
  pages = {110539},
  year = {2023},
  issn = {0950-7051},
  doi = {10.1016/j.knosys.2023.110539},
  author = {Carlos Gonz{\'a}lez-Santos and Miguel A. Vega-Rodr{\'\i}guez and Carlos J. P{\'e}rez and Joaqu{\'\i}n M. L{\'o}pez-Mu{\~n}oz and I{\~n}aki Mart{\'\i}nez-Sarriegui}
}

@inproceedings{Kiesel2022,
  title = "Identifying the Human Values behind Arguments",
  author = "Kiesel, Johannes  and Alshomary, Milad  and Handke, Nicolas  and Cai, Xiaoni  and Wachsmuth, Henning  and Stein, Benno",
  editor = "Muresan, Smaranda  and Nakov, Preslav  and Villavicencio, Aline",
  booktitle = "Proceedings of the 60th Annual Meeting of the Association for Computational Linguistics (Volume 1: Long Papers)",
  month = may,
  year = "2022",
  address = "Dublin, Ireland",
  publisher = "Association for Computational Linguistics",
  doi = "10.18653/v1/2022.acl-long.306",
  pages = "4459--4471",
}

@inproceedings{Mirzakhmedova2024,
  title = "The Touch{\'e}23-{V}alue{E}val Dataset for Identifying Human Values behind Arguments",
  author = "Mirzakhmedova, Nailia  and Kiesel, Johannes  and Alshomary, Milad  and Heinrich, Maximilian  and Handke, Nicolas  and Cai, Xiaoni  and Barriere, Valentin  and Dastgheib, Doratossadat  and Ghahroodi, Omid  and SadraeiJavaheri, MohammadAli  and Asgari, Ehsaneddin  and Kawaletz, Lea  and Wachsmuth, Henning  and Stein, Benno",
  editor = "Calzolari, Nicoletta  and Kan, Min-Yen  and Hoste, Veronique  and Lenci, Alessandro  and Sakti, Sakriani  and Xue, Nianwen",
  booktitle = "Proceedings of the 2024 Joint International Conference on Computational Linguistics, Language Resources and Evaluation (LREC-COLING 2024)",
  month = may,
  year = "2024",
  address = "Torino, Italia",
  publisher = "ELRA and ICCL",
  url = "https://aclanthology.org/2024.lrec-main.1402/",
  pages = "16121--16134",
}

@inproceedings{Yang2016,
  title={Hierarchical attention networks for document classification},
  author={Yang, Zichao and Yang, Diyi and Dyer, Chris and He, Xiaodong and Smola, Alex and Hovy, Eduard},
  booktitle={Proceedings of the 2016 conference of the North American chapter of the association for computational linguistics: human language technologies},
  pages={1480--1489},
  year={2016}
}

@inproceedings{Devlin2019,
    title = "{BERT}: Pre-training of Deep Bidirectional Transformers for Language Understanding",
    author = "Devlin, Jacob  and Chang, Ming-Wei  and Lee, Kenton  and Toutanova, Kristina",
    editor = "Burstein, Jill  and Doran, Christy  and Solorio, Thamar",
    booktitle = "Proceedings of the 2019 Conference of the North {A}merican Chapter of the Association for Computational Linguistics: Human Language Technologies, Volume 1 (Long and Short Papers)",
    month = jun,
    year = "2019",
    address = "Minneapolis, Minnesota",
    publisher = "Association for Computational Linguistics",
    doi = "10.18653/v1/N19-1423",
    pages = "4171--4186",
}

@InProceedings{Yeste2024,
  author    = {V{\'\i}ctor Yeste and Mariona Coll-Ardanuy and Paolo Rosso},
  title     = {Philo of Alexandria at Touch{\'e}: A Cascade Model Approach to Human Value Detection},
  booktitle = {Working Notes Papers of the CLEF 2024 Evaluation Labs},
  editor    = {Guglielmo Faggioli and Nicola Ferro and Petra Galuscakova and Alba Garc{\'i}a Seco Herrera},
  series    = {CEUR Workshop Proceedings},
  volume    = {3740},
  pages     = {3503--3508},
  month     = sep,
  year      = {2024},
  site      = {Grenoble, France},
  url       = {http://ceur-ws.org/Vol-3740/paper-338.pdf},
}

@article{Graham2011,
  author  = {Jesse Graham and Brian A. Nosek and Jonathan Haidt and Ravi Iyer and Spassena Koleva and Peter H. Ditto},
  title   = {Mapping the Moral Domain},
  journal = {Journal of Personality and Social Psychology},
  volume  = {101},
  number  = {2},
  pages   = {366--385},
  year    = {2011},
  doi     = {10.1037/a0021847}
}

@book{Haidt2012,
  author    = {Haidt, Jonathan},
  title     = {The Righteous Mind: Why Good People Are Divided by Politics and Religion},
  publisher = {Pantheon Books},
  year      = {2012},
  isbn      = {978-0-307-37790-6}
}

@article{Hopp2021,
  author="Hopp, Frederic R. and Fisher, Jacob T. and Cornell, Devin and Huskey, Richard and Weber, Ren{\'e}",
  title="The extended Moral Foundations Dictionary (eMFD): Development and applications of a crowd-sourced approach to extracting moral intuitions from text",
  journal="Behavior Research Methods",
  year="2021",
  month="Feb",
  day="01",
  volume="53",
  number="1",
  pages="232--246",
  issn="1554-3528",
  doi="10.3758/s13428-020-01433-0",
}

@article{Hoover2020,
  author = {Joe Hoover and Gwenyth Portillo-Wightman and Leigh Yeh and Shreya Havaldar and Aida Mostafazadeh Davani and Ying Lin and Brendan Kennedy and Mohammad Atari and Zahra Kamel and Madelyn Mendlen and Gabriela Moreno and Christina Park and Tingyee E. Chang and Jenna Chin and Christian Leong and Jun Yen Leung and Arineh Mirinjian and Morteza Dehghani},
  title ={Moral Foundations Twitter Corpus: A Collection of 35k Tweets Annotated for Moral Sentiment},
  journal = {Social Psychological and Personality Science},
  volume = {11},
  number = {8},
  pages = {1057-1071},
  year = {2020},
  doi = {10.1177/1948550619876629},
}

@Article{Breiman1996,
  author="Breiman, Leo",
  title="Bagging predictors",
  journal="Machine Learning",
  year="1996",
  month="Aug",
  day="01",
  volume="24",
  number="2",
  pages="123--140",
  issn="1573-0565",
  doi="10.1007/BF00058655",
}

@Article{Breiman2001,
  author="Breiman, Leo",
  title="Random Forests",
  journal="Machine Learning",
  year="2001",
  month="Oct",
  day="01",
  volume="45",
  number="1",
  pages="5--32",
  issn="1573-0565",
  doi="10.1023/A:1010933404324",
}

@article{FreundSchapire1997,
  title = {A Decision-Theoretic Generalization of On-Line Learning and an Application to Boosting},
  journal = {Journal of Computer and System Sciences},
  volume = {55},
  number = {1},
  pages = {119-139},
  year = {1997},
  issn = {0022-0000},
  doi = {10.1006/jcss.1997.1504},
  author = {Yoav Freund and Robert E Schapire},
}

@article{Wolpert1992,
  title = {Stacked generalization},
  journal = {Neural Networks},
  volume = {5},
  number = {2},
  pages = {241-259},
  year = {1992},
  issn = {0893-6080},
  doi = {10.1016/S0893-6080(05)80023-1},
  author = {David H. Wolpert},
}

@InProceedings{Zhao2021,
  title = 	 {Calibrate Before Use: Improving Few-shot Performance of Language Models},
  author =       {Zhao, Zihao and Wallace, Eric and Feng, Shi and Klein, Dan and Singh, Sameer},
  booktitle = 	 {Proceedings of the 38th International Conference on Machine Learning},
  pages = 	 {12697--12706},
  year = 	 {2021},
  editor = 	 {Meila, Marina and Zhang, Tong},
  volume = 	 {139},
  series = 	 {Proceedings of Machine Learning Research},
  month = 	 {18--24 Jul},
  publisher =    {PMLR},
  url = 	 {https://proceedings.mlr.press/v139/zhao21c.html},
}

@inproceedings{Wei2022,
  author = {Wei, Jason and Wang, Xuezhi and Schuurmans, Dale and Bosma, Maarten and ichter, brian and Xia, Fei and Chi, Ed and Le, Quoc V and Zhou, Denny},
  booktitle = {Advances in Neural Information Processing Systems},
  editor = {S. Koyejo and S. Mohamed and A. Agarwal and D. Belgrave and K. Cho and A. Oh},
  pages = {24824--24837},
  publisher = {Curran Associates, Inc.},
  title = {Chain-of-Thought Prompting Elicits Reasoning in Large Language Models},
  url = {https://proceedings.neurips.cc/paper\_files/paper/2022/file/9d5609613524ecf4f15af0f7b31abca4-Paper-Conference.pdf},
  volume = {35},
  year = {2022}
}

@article{Liu2023,
  author = {Liu, Pengfei and Yuan, Weizhe and Fu, Jinlan and Jiang, Zhengbao and Hayashi, Hiroaki and Neubig, Graham},
  title = {Pre-train, Prompt, and Predict: A Systematic Survey of Prompting Methods in Natural Language Processing},
  year = {2023},
  issue_date = {September 2023},
  publisher = {Association for Computing Machinery},
  address = {New York, NY, USA},
  volume = {55},
  number = {9},
  issn = {0360-0300},
  doi = {10.1145/3560815},
  journal = {ACM Comput. Surv.},
  month = jan,
  articleno = {195},
  numpages = {35},
}

@Article{SilvaFilho2023,
  author="Silva Filho, Telmo and Song, Hao and Perello-Nieto, Miquel and Santos-Rodriguez, Raul and Kull, Meelis and Flach, Peter",
  title="Classifier calibration: a survey on how to assess and improve predicted class probabilities",
  journal="Machine Learning",
  year="2023",
  month="Sep",
  day="01",
  volume="112",
  number="9",
  pages="3211--3260",
  issn="1573-0565",
  doi="10.1007/s10994-023-06336-7"
}

@inproceedings{Valmadre2022,
  author = {Valmadre, Jack},
  booktitle = {Advances in Neural Information Processing Systems},
  editor = {S. Koyejo and S. Mohamed and A. Agarwal and D. Belgrave and K. Cho and A. Oh},
  pages = {18034--18045},
  publisher = {Curran Associates, Inc.},
  title = {Hierarchical classification at multiple operating points},
  url = {https://proceedings.neurips.cc/paper\_files/paper/2022/file/727855c31df8821fd18d41c23daebf10-Paper-Conference.pdf},
  volume = {35},
  year = {2022}
}

@inproceedings{Rink2025,
  author="Rink, Olga and Maysuradze, Archil and Fedorov, Artem and Ischenko, Roman and Korchagina, Anna and Tabachenkov, Andrey and Tsybanov, Ilya and Vorontsov, Konstantin",
  editor="Coman, Adela
  and Vasilache, Simona",
  title="Automated Detection of Human Values in Texts: ML Challenges and Performance Benchmarks",
  booktitle="Social Computing and Social Media",
  year="2025",
  publisher="Springer Nature Switzerland",
  address="Cham",
  pages="304--321",
  isbn="978-3-031-93536-7"
}

@misc{Sun2024,
  title={Fine-tuning vs Prompting, Can Language Models Understand Human Values?}, 
  author={Pingwei Sun},
  year={2024},
  eprint={2403.09720},
  archivePrefix={arXiv},
  primaryClass={cs.CL},
  url={https://arxiv.org/abs/2403.09720}, 
}

@InProceedings{GarciaRodriguez2025,
  author="Garc{\'i}a-Rodr{\'i}guez, Sara and Karanik, Marcelo and Pina-Zapata, Alicia",
  editor="Osman, Nardine
  and Steels, Luc",
  title="Value Promotion Scheme Elicitation Using Natural Language Processing: A Model for Value-Based Agent Architecture",
  booktitle="Value Engineering in Artificial Intelligence",
  year="2025",
  publisher="Springer Nature Switzerland",
  address="Cham",
  pages="104--120",
  isbn="978-3-031-85463-7"
}

@inproceedings{Reinig2024,
  title = "A Survey on Modelling Morality for Text Analysis",
  author = "Reinig, Ines  and Becker, Maria  and Rehbein, Ines  and Ponzetto, Simone",
  editor = "Ku, Lun-Wei  and Martins, Andre  and Srikumar, Vivek",
  booktitle = "Findings of the Association for Computational Linguistics: ACL 2024",
  month = aug,
  year = "2024",
  address = "Bangkok, Thailand",
  publisher = "Association for Computational Linguistics",
  doi = "10.18653/v1/2024.findings-acl.245",
  pages = "4136--4155",
}

@misc{Boyd2022,
  title        = {The Development and Psychometric Properties of {LIWC}-22},
  author       = {Boyd, Ryan L. and Ashokkumar, Abhishek and Seraj, Sarah and Pennebaker, James W.},
  year         = {2022},
  howpublished = {Technical report / manual, LIWC},
  note         = {LIWC-22 documentation},
}

@article{MohammadTurney2013,
  author = {Mohammad, Saif M. and Turney, Peter D.},
  title = {Crowdsourcing a Word--Emotion Association Lexicon},
  journal = {Computational Intelligence},
  volume = {29},
  number = {3},
  pages = {436-465},
  doi = {10.1111/j.1467-8640.2012.00460.x},
  year = {2013}
}

@inproceedings{MohammadKiritchenko2018,
  title     = {Understanding emotions: A dataset of tweets to study interactions between affect categories},
  author    = {Mohammad, Saif M. and Kiritchenko, Svetlana},
  booktitle = {Proceedings of the 11th International Conference on Language Resources and Evaluation (LREC 2018)},
  year      = {2018},
}

@Article{Warriner2013,
  author="Warriner, Amy Beth and Kuperman, Victor and Brysbaert, Marc",
  title="Norms of valence, arousal, and dominance for 13,915 English lemmas",
  journal="Behavior Research Methods",
  year="2013",
  month="Dec",
  day="01",
  volume="45",
  number="4",
  pages="1191--1207",
  issn="1554-3528",
  doi="10.3758/s13428-012-0314-x",
}

@inproceedings{Mohammad2018,
    title = "Obtaining Reliable Human Ratings of Valence, Arousal, and Dominance for 20,000 {E}nglish Words",
    author = "Mohammad, Saif",
    editor = "Gurevych, Iryna  and Miyao, Yusuke",
    booktitle = "Proceedings of the 56th Annual Meeting of the Association for Computational Linguistics (Volume 1: Long Papers)",
    month = jul,
    year = "2018",
    address = "Melbourne, Australia",
    publisher = "Association for Computational Linguistics",
    doi = "10.18653/v1/P18-1017",
    pages = "174--184",
}

@inproceedings{Mohammad2024,
  title = "{W}orry{W}ords: Norms of Anxiety Association for over 44k {E}nglish Words",
  author = "Mohammad, Saif M.",
  editor = "Al-Onaizan, Yaser  and Bansal, Mohit  and Chen, Yun-Nung",
  booktitle = "Proceedings of the 2024 Conference on Empirical Methods in Natural Language Processing",
  month = nov,
  year = "2024",
  address = "Miami, Florida, USA",
  publisher = "Association for Computational Linguistics",
  doi = "10.18653/v1/2024.emnlp-main.910",
  pages = "16261--16278",
}

@article{Blei200,
  title   = {Latent Dirichlet Allocation},
  author  = {Blei, David M. and Ng, Andrew Y. and Jordan, Michael I.},
  journal = {Journal of Machine Learning Research},
  year    = {2003},
  volume  = {3},
  pages   = {993--1022},
}

@Article{LeeSeung1999,
  author="Lee, Daniel D. and Seung, H. Sebastian",
  title="Learning the parts of objects by non-negative matrix factorization",
  journal="Nature",
  year="1999",
  month="Oct",
  day="01",
  volume="401",
  number="6755",
  pages="788--791",
  issn="1476-4687",
  doi="10.1038/44565",
}

@misc{Grootendorst2022,
  title={BERTopic: Neural topic modeling with a class-based TF-IDF procedure}, 
  author={Maarten Grootendorst},
  year={2022},
  eprint={2203.05794},
  archivePrefix={arXiv},
  primaryClass={cs.CL},
  url={https://arxiv.org/abs/2203.05794}, 
}

@Inbook{Platt1999,
  title     = {Probabilistic Outputs for Support Vector Machines and Comparisons to Regularized Likelihood Methods},
  author    = {Platt, John},
  booktitle = {Advances in Large Margin Classifiers},
  editor    = {Smola, Alex J. and Bartlett, Peter and Sch{\"o}lkopf, Bernhard and Schuurmans, Dale},
  volume    = {20},
  chapter   = {3},
  pages     = {61--74},
  year      = {1999},
  publisher = {MIT Press}
}

@inproceedings{Borenstein2025,
  title = "Investigating Human Values in Online Communities",
  author = "Borenstein, Nadav  and Arora, Arnav  and Kaffee, Lucie-Aim{\'e}e  and Augenstein, Isabelle",
  editor = "Chiruzzo, Luis  and Ritter, Alan  and Wang, Lu",
  booktitle = "Proceedings of the 2025 Conference of the Nations of the Americas Chapter of the Association for Computational Linguistics: Human Language Technologies (Volume 1: Long Papers)",
  month = apr,
  year = "2025",
  address = "Albuquerque, New Mexico",
  publisher = "Association for Computational Linguistics",
  doi = "10.18653/v1/2025.naacl-long.77",
  pages = "1607--1627",
  isbn = "979-8-89176-189-6",
}

@misc{Starovolsky-Shitrit2025,
  title={The Value of Nothing: Multimodal Extraction of Human Values Expressed by TikTok Influencers}, 
  author={Alina Starovolsky-Shitrit and Alon Neduva and Naama Appel Doron and Ella Daniel and Oren Tsur},
  year={2025},
  eprint={2501.11770},
  archivePrefix={arXiv},
  primaryClass={cs.CL},
  url={https://arxiv.org/abs/2501.11770}, 
}

@misc{Segerer2025,
  title={Cultural Value Alignment in Large Language Models: A Prompt-based Analysis of Schwartz Values in Gemini, ChatGPT, and DeepSeek}, 
  author={Robin Segerer},
  year={2025},
  eprint={2505.17112},
  archivePrefix={arXiv},
  primaryClass={cs.CL},
  url={https://arxiv.org/abs/2505.17112}, 
}

@InProceedings{Rink2024,
  author="Rink, Olga and Lobachev, Viktor and Vorontsov, Konstantin",
  editor="Coman, Adela and Vasilache, Simona",
  title="Detecting Human Values and Sentiments in Large Text Collections with a Context-Dependent Information Markup: A Methodology and Math",
  booktitle="Social Computing and Social Media",
  year="2024",
  publisher="Springer Nature Switzerland",
  address="Cham",
  pages="372--383",
  isbn="978-3-031-61281-7"
}

@inproceedings{Yao2024,
  title = "Value {FULCRA}: Mapping Large Language Models to the Multidimensional Spectrum of Basic Human Value",
  author = "Yao, Jing  and Yi, Xiaoyuan  and Gong, Yifan  and Wang, Xiting  and Xie, Xing",
  editor = "Duh, Kevin  and Gomez, Helena  and Bethard, Steven",
  booktitle = "Proceedings of the 2024 Conference of the North American Chapter of the Association for Computational Linguistics: Human Language Technologies (Volume 1: Long Papers)",
  month = jun,
  year = "2024",
  address = "Mexico City, Mexico",
  publisher = "Association for Computational Linguistics",
  doi = "10.18653/v1/2024.naacl-long.486",
  pages = "8762--8785",
}

@article{Biedma2024,
  publtype={informal},
  author={Pablo Biedma and Xiaoyuan Yi and Linus Huang and Maosong Sun and Xing Xie},
  title={Beyond Human Norms: Unveiling Unique Values of Large Language Models through Interdisciplinary Approaches},
  year={2024},
  cdate={1704067200000},
  journal={CoRR},
  volume={abs/2404.12744},
  url={10.48550/arXiv.2404.12744}
}

@misc{Zhu2025,
  title={EAVIT: Efficient and Accurate Human Value Identification from Text data via LLMs}, 
  author={Wenhao Zhu and Yuhang Xie and Guojie Song and Xin Zhang},
  year={2025},
  eprint={2505.12792},
  archivePrefix={arXiv},
  primaryClass={cs.CL},
  url={https://arxiv.org/abs/2505.12792}, 
}

@inproceedings{Shen2025,
  title = "{V}alue{C}ompass: A Framework for Measuring Contextual Value Alignment Between Human and {LLM}s",
  author = "Shen, Hua  and Knearem, Tiffany  and Ghosh, Reshmi  and Yang, Yu-Ju  and Clark, Nicholas  and Mitra, Tanu  and Huang, Yun",
  editor = "Zhang, Chen  and Allaway, Emily  and Shen, Hua  and Miculicich, Lesly  and Li, Yinqiao  and M'hamdi, Meryem  and Limkonchotiwat, Peerat  and Bai, Richard He  and T.y.s.s., Santosh  and Han, Sophia Simeng  and Thapa, Surendrabikram  and Rim, Wiem Ben",
  booktitle = "Proceedings of the 9th Widening NLP Workshop",
  month = nov,
  year = "2025",
  address = "Suzhou, China",
  publisher = "Association for Computational Linguistics",
  doi = "10.18653/v1/2025.winlp-main.15",
  pages = "75--86",
  isbn = "979-8-89176-351-7",
}

@article{Ye2025,
  title={Measuring Human and AI Values Based on Generative Psychometrics with Large Language Models},
  volume={39},
  doi={10.1609/aaai.v39i25.34839},
  number={25},
  journal={Proceedings of the AAAI Conference on Artificial Intelligence},
  author={Ye, Haoran and Xie, Yuhang and Ren, Yuanyi and Fang, Hanjun and Zhang, Xin and Song, Guojie},
  year={2025},
  month={Apr.},
  pages={26400-26408}
}

@inproceedings{Rozen2025,
  title={Do {LLM}s have Consistent Values?},
  author={Naama Rozen and Liat Bezalel and Gal Elidan and Amir Globerson and Ella Daniel},
  booktitle={The Thirteenth International Conference on Learning Representations},
  year={2025},
  url={https://openreview.net/forum?id=8zxGruuzr9}
}

@inproceedings{Chen2025,
  title = "{M}o{V}a: Towards Generalizable Classification of Human Morals and Values",
  author = "Chen, Ziyu  and Sun, Junfei  and Li, Chenxi  and Nguyen, Tuan Dung  and Yao, Jing  and Yi, Xiaoyuan  and Xie, Xing  and Tan, Chenhao  and Xie, Lexing",
  editor = "Christodoulopoulos, Christos  and Chakraborty, Tanmoy  and Rose, Carolyn  and Peng, Violet",
  booktitle = "Proceedings of the 2025 Conference on Empirical Methods in Natural Language Processing",
  month = nov,
  year = "2025",
  address = "Suzhou, China",
  publisher = "Association for Computational Linguistics",
  doi = "10.18653/v1/2025.emnlp-main.1687",
  pages = "33216--33260",
  isbn = "979-8-89176-332-6",
}

@inproceedings{Cahyawijaya2025,
  title = "High-Dimension Human Value Representation in Large Language Models",
  author = "Cahyawijaya, Samuel  and Chen, Delong  and Bang, Yejin  and Khalatbari, Leila  and Wilie, Bryan  and Ji, Ziwei  and Ishii, Etsuko  and Fung, Pascale",
  editor = "Chiruzzo, Luis  and Ritter, Alan  and Wang, Lu",
  booktitle = "Proceedings of the 2025 Conference of the Nations of the Americas Chapter of the Association for Computational Linguistics: Human Language Technologies (Volume 1: Long Papers)",
  month = apr,
  year = "2025",
  address = "Albuquerque, New Mexico",
  publisher = "Association for Computational Linguistics",
  doi = "10.18653/v1/2025.naacl-long.274",
  pages = "5303--5330",
  ISBN = "979-8-89176-189-6",
}

@article{Wang2024,
  title = {Hierarchical classification with exponential weighting of multi-granularity paths},
  author = {Yibin Wang and Qing Zhu and Yusheng Cheng},
  journal = {Information Sciences},
  volume = {675},
  pages = {120715},
  year = {2024},
  issn = {0020-0255},
  doi = {10.1016/j.ins.2024.120715},
}

@article{Zhang2022,
  title = {LA-HCN: Label-based Attention for Hierarchical Multi-label Text Classification Neural Network},
  author = {Xinyi Zhang and Jiahao Xu and Charlie Soh and Lihui Chen},
  journal = {Expert Systems with Applications},
  volume = {187},
  pages = {115922},
  year = {2022},
  issn = {0957-4174},
  doi = {10.1016/j.eswa.2021.115922},
}

@article{Liu2024,
  title = {Improve label embedding quality through global sensitive GAT for hierarchical text classification},
  author = {Hankai Liu and Xianying Huang and Xiaoyang Liu},
  journal = {Expert Systems with Applications},
  volume = {238},
  pages = {122267},
  year = {2024},
  issn = {0957-4174},
  doi = {10.1016/j.eswa.2023.122267},
}

@article{Nunez2026,
  title = {Design of a GenAI UX layer with small language models for edge computing in smart agriculture},
  journal = {Array},
  volume = {29},
  pages = {100632},
  year = {2026},
  issn = {2590-0056},
  doi = {10.1016/j.array.2025.100632},
  author = {Juan M. {Núñez V.} and Carlos Alberto Peláez and Andrés Solano and Juan M. Corchado and Fernando {De la Prieta}},
}

@article{Zhou2026,
  title = {Automated quality assessment of online medical science popularization Videos: A framework based on large language models},
  journal = {Array},
  volume = {29},
  pages = {100717},
  year = {2026},
  issn = {2590-0056},
  doi = {10.1016/j.array.2026.100717},
  author = {Shiqi Zhou and Hua Wu and Yongjian Zhang and Qianqian Zhong and Yueqin Diao and Huihui Fang and Yanwu Xu and Hanyi Yu},
}

@article{Zhu2023,
  title = {Dynamic ensemble learning for multi-label classification},
  author = {Xiaoyan Zhu and Jiaxuan Li and Jingtao Ren and Jiayin Wang and Guangtao Wang},
  journal = {Information Sciences},
  volume = {623},
  pages = {94-111},
  year = {2023},
  issn = {0020-0255},
  doi = {10.1016/j.ins.2022.12.022},
}

@article{Stein2019,
  title = {An analysis of hierarchical text classification using word embeddings},
  author = {Roger Alan Stein and Patricia A. Jaques and João Francisco Valiati},
  journal = {Information Sciences},
  volume = {471},
  pages = {216-232},
  year = {2019},
  issn = {0020-0255},
  doi = {10.1016/j.ins.2018.09.001},
}

\end{document}